\documentclass{article}

     \usepackage[final,nonatbib]{neurips_2022}

\usepackage[utf8]{inputenc} 
\usepackage[T1]{fontenc}    
\usepackage{hyperref}       
\usepackage{url}            
\usepackage{booktabs}       
\usepackage{amsfonts}       
\usepackage{nicefrac}       
\usepackage{microtype}      
\usepackage{xcolor}         
\usepackage{amsmath}
\usepackage{graphicx}
\usepackage{subfigure}
\usepackage{amsthm,amsmath,amssymb}
\usepackage{mathrsfs}
\usepackage{booktabs}
\usepackage{tablefootnote}
\usepackage{threeparttable}
\usepackage{bm}
\usepackage{mathtools}
\usepackage{multirow}
\usepackage{caption}
\usepackage{wrapfig}
\usepackage[noend]{algpseudocode}
\usepackage{algorithmicx,algorithm}

\title{Online Training Through Time for Spiking Neural Networks}

\author{%
  Mingqing Xiao$^1$, Qingyan Meng$^{2,3}$, Zongpeng Zhang$^{4}$, Di He$^{1}$, Zhouchen Lin$^{1,5,6\thanks{Corresponding author.}}$ \\
  $^1$Key Lab. of Machine Perception (MoE), School of Intelligence Science and Technology, \\Peking University\\
  $^2$The Chinese University of Hong Kong, Shenzhen\\
  $^3$Shenzhen Research Institute of Big Data\\
  $^4$Center for Data Science, Academy for Advanced Interdisciplinary Studies, Peking University\\
  $^5$Institute for Artificial Intelligence, Peking University\\
  $^6$Peng Cheng Laboratory, China\\
  \texttt{\{mingqing\_xiao, dihe, zlin\}@pku.edu.cn, qingyanmeng@link.cuhk.edu.cn,}\\ \texttt{zongpeng.zhang98@gmail.com} \\
}

\begin{document}

\maketitle

\begin{abstract}
  Spiking neural networks (SNNs) are promising brain-inspired energy-efficient models. Recent progress in training methods has enabled successful deep SNNs on large-scale tasks with low latency. Particularly, backpropagation through time (BPTT) with surrogate gradients (SG) is popularly used to enable models to achieve high performance in a very small number of time steps. However, it is at the cost of large memory consumption for training, lack of theoretical clarity for optimization, and inconsistency with the online property of biological learning rules and rules on neuromorphic hardware. Other works connect the spike representations of SNNs with equivalent artificial neural network formulation and train SNNs by gradients from equivalent mappings to ensure descent directions. But they fail to achieve low latency and are also not online. In this work, we propose online training through time (OTTT) for SNNs, which is derived from BPTT to enable forward-in-time learning by tracking presynaptic activities and leveraging instantaneous loss and gradients. Meanwhile, we theoretically analyze and prove that the gradients of OTTT can provide a similar descent direction for optimization as gradients from equivalent mapping between spike representations under both feedforward and recurrent conditions. OTTT only requires constant training memory costs agnostic to time steps, avoiding the significant memory costs of BPTT for GPU training. Furthermore, the update rule of OTTT is in the form of three-factor Hebbian learning, which could pave a path for online on-chip learning. With OTTT, it is the first time that the two mainstream supervised SNN training methods, BPTT with SG and spike representation-based training, are connected, and meanwhile it is in a biologically plausible form. Experiments on CIFAR-10, CIFAR-100, ImageNet, and CIFAR10-DVS demonstrate the superior performance of our method on large-scale static and neuromorphic datasets in a small number of time steps. Our code is available at \url{https://github.com/pkuxmq/OTTT-SNN}.
\end{abstract}

\vspace{-3mm}
\section{Introduction}
\vspace{-2mm}
Spiking neural networks (SNNs) are regarded as the third generation of neural network models~\cite{maass1997networks} and have gained increasing attention in recent years~\cite{lee2016training,shrestha2018slayer,wu2018spatio,roy2019towards,zheng2020going,deng2021optimal,li2021free,wu2021tandem,li2021differentiable,fang2021deep,xiao2021training,deng2021temporal}. SNNs are composed of brain-inspired spiking neurons that imitate biological neurons to transmit spikes between each other. This allows event-based computation and enables efficient computation on neuromorphic hardware with low energy consumption~\cite{akopyan2015truenorth,davies2018loihi,pei2019towards}. 

However, the supervised training of SNNs is challenging due to the non-differentiable neuron model with discrete spike-generation procedures. Several kinds of methods are proposed to tackle the problem, and recent progress has empirically obtained successful results. Backpropagation through time (BPTT) with surrogate gradients (SG) is one of the mainstream methods which enables the training of deep SNNs with high performance on large-scale datasets (e.g., ImageNet) with extremely low latency (e.g., 4-6 time steps)~\cite{zheng2020going,li2021differentiable,fang2021deep,deng2021temporal}. These approaches unfold the iterative expression of spiking neurons, backpropagate the errors through time~\cite{werbos1990backpropagation}, and use surrogate derivatives to approximate the gradient of the spiking function ~\cite{shrestha2018slayer,wu2018spatio,bellec2018long,jin2018hybrid,wu2019direct,neftci2019surrogate,kim2020unifying,Fang_2021_ICCV}. As a result, during training, they suffer from significant memory costs proportional to the number of time steps, and the optimization with approximated surrogate gradients is not well guaranteed theoretically. 
Another branch of works builds the closed-form formulation for the spike representation of neurons, e.g. the (weighted) firing rate or spiking time, which is similar to conventional artificial neural networks (ANNs). Then SNNs can be either optimized by calculating the gradients from the equivalent mappings between spike representations~\cite{lee2016training,thiele2019spikegrad,wu2021training,zhou2021temporal,wu2021tandem,meng2022training}, or converted from a trained equivalent ANN counterpart~\cite{hunsberger2015spiking,rueckauer2017conversion,sengupta2019going,rathi2019enabling,han2020deep,deng2021optimal,yan2021near,li2021free,stockl2021optimized}. The optimization of these methods is clearer than surrogate gradients. However, they require a larger number of time steps compared to BPTT with SG. Therefore, they suffer from high latency, and more energy consumption is required if the representation is rate-based. Another critical point for both methods is that they are indeed inconsistent with biological online learning, which is also the learning rule on neuromorphic hardware~\cite{davies2018loihi}. 

In this work, we develop a novel approach for training SNNs to achieve high performance with low latency, and maintain the online learning property to pave a path for learning on neuromorphic chips. We call our method online training through time (OTTT). We first derive OTTT from the commonly used BPTT with SG method by analyzing the temporal dependency and proposing to track the presynaptic activities in order to decouple this dependency. With the instantaneous loss calculation, OTTT can perform forward-in-time learning, i.e. calculations are done online in time without computing backward through the time. Then we theoretically analyze the gradients of OTTT and gradients of spike representation-based methods. 
We show that they have similar expressions and prove that they can provide the similar descent direction for the optimization problem formulated by spike representation. For the feedforward network condition, gradients are easily calculated and analyzed. For the recurrent network condition, we follow the framework in~\cite{xiao2021training} that weighted firing rates will converge to an equilibrium state and gradients can be calculated by implicit differentiation. With this formulation, the gradients correspond to an approximation of gradients calculated by implicit differentiation, which can be proved to be a descent direction for the optimization problem as well~\cite{fung2021jfb,geng2021training}. 
In this way, a connection between OTTT and spike representation-based methods is bridged. Finally, we show that OTTT is in the form of three-factor Hebbian learning rule~\cite{fremaux2016neuromodulated}, which could pave a path for online learning on neuromorphic chips. Our contributions include:

\vspace{-2mm}
\begin{enumerate}
    \item We propose online training through time (OTTT) for SNNs, which enables forward-in-time learning and only requires constant training memory agnostic to time steps, avoiding the large training memory costs of backpropagation through time (BPTT).
    \item We theoretically analyze and connect the gradients of OTTT and gradients based on spike representations, and prove the descent guarantee for optimization under both feedforward and recurrent conditions.
    \item We show that OTTT is in the form of three-factor Hebbian learning rule, which could pave a path for on-chip online learning. With OTTT, it is the first time that a connection between BPTT with SG, spike representation-based methods, and biological three-factor Hebbian learning is bridged.
    \item We conduct extensive experiments on CIFAR-10, CIFAR-100, ImageNet, and CIFAR10-DVS, which demonstrate the superior results of our methods on large-scale static and neuromorphic datasets in a small number of time steps.
\end{enumerate}

\vspace{-2mm}
\section{Related Work}
\vspace{-2mm}

\textbf{SNN Training Methods.}\quad As for supervised training of SNNs, there are two main research directions. One direction is to build a connection between spike representations (e.g. firing rates) of SNNs with equivalent ANN-like closed-form mappings. With the connection, SNNs can be converted from ANNs~\cite{hunsberger2015spiking,rueckauer2017conversion,sengupta2019going,rathi2019enabling,han2020deep,deng2021optimal,yan2021near,li2021free,stockl2021optimized,meng2022trainingnn}, or SNNs can be optimized by gradients calculated from equivalent mappings~\cite{lee2016training,thiele2019spikegrad,wu2021training,zhou2021temporal,wu2021tandem,meng2022training}. Variants following this direction also include~\cite{xiao2021training} which connects feedback SNNs with equilibrium states following fixed-point equations instead of closed-form mappings. These methods have a clearer descent direction for the optimization problem, but require a relatively large number of time steps, suffering from high latency and usually more energy consumption with rate based representation. Another direction is to directly calculate gradients based on the SNN computation. They follow the BPTT framework, and deal with the non-differentiable problem of spiking functions by applying surrogate gradients~\cite{shrestha2018slayer,wu2018spatio,bellec2018long,jin2018hybrid,wu2019direct,neftci2019surrogate,zheng2020going,Fang_2021_ICCV,li2021differentiable,fang2021deep,deng2021temporal}, or computing gradients with respect to spiking times based on the linear assumption~\cite{bohte2002error,zhang2020temporal}, or combining both~\cite{kim2020unifying}. BPTT with SG can achieve extremely low latency. However, it requires large training memory to maintain the computational graph unfolded along time, and it remains unknown why surrogate gradients work well. \cite{li2021differentiable} empirically observed that surrogate gradients have a high similarity with numerical gradients, but it remains unclear theoretically. And gradients based on spiking times suffer from the ``dead neuron'' problem~\cite{shrestha2018slayer}, so they should be combined with SG in practice~\cite{zhang2020temporal,kim2020unifying}. 
Meanwhile, methods in both directions are inconsistent with biological online learning, i.e. forward-in-time learning, to pave a path for learning on neuromorphic hardware. Differently, our proposed method avoids the above problems and maintain the online property.

\textbf{Online Training of Neural Networks.}\quad In the research of recurrent neural networks (RNNs), there are several alternatives for BPTT to enable online learning. Particularly, real time recurrent learning (RTRL)~\cite{williams1989learning} proposes to propagate partial derivatives of hidden states over parameters through time to enable forward-in-time calculation of gradients. Several recent works improve the memory costs of RTRL with approximation for more practical usage~\cite{tallec2018unbiased,mujika2018approximating,menick2020practical}. Another work proposes to online update parameters based on decoupled gradients with regularization at each time step~\cite{kag2021training}. However, these are all for RNNs and not tailored to SNNs. Several online training methods are proposed for SNNs~\cite{zenke2018superspike,bellec2020solution,bohnstingl2022online}, which are derived in the spirit of RTRL and simplified for SNNs. \cite{kaiser2020synaptic} leverages local losses and ignores temporal dependencies for online local training of SNNs, and \cite{yin2021accurate} directly apply the method in \cite{kag2021training} to train SNNs. However, these methods also leverage surrogate gradients without providing theoretical explanation for optimization. Meanwhile, \cite{zenke2018superspike,kaiser2020synaptic} use feedback alignment~\cite{nokland2016direct}, \cite{bellec2020solution} is limited to single-layer recurrent SNNs, and \cite{bohnstingl2022online} requires much larger memory costs for eligibility traces, so they cannot scale to large-scale tasks. \cite{yin2021accurate} requires a specially designed neuron model and more computation for parameter regularization, and also does not consider large tasks. 
Differently, our work explain the descent direction under both feedforward and recurrent conditions with convergent inputs, and is efficient and scalable to large-scale tasks including ImageNet classification.

\vspace{-1mm}
\section{Preliminaries}
\vspace{-2mm}

\subsection{Spiking Neural Networks}
\vspace{-2mm}

Spiking neurons are brain-inspired models that transmit information by spike trains. Each neuron maintains a membrane potential $u$ and integrates input spike trains, which will generate a spike once $u$ exceeds a threshold. We consider the commonly used leaky integrate and fire (LIF) model, which describes the dynamics of the membrane potential as:
\vspace{-1mm}
\begin{equation}\label{eq.origin}
    \tau_m\frac{du}{dt} = -(u-u_{rest}) + R\cdot I(t),\quad u < V_{th},
\end{equation}
where $I$ is the input current, $V_{th}$ is the threshold, and $R$ and $\tau_m$ are resistance and time constant, respectively. 
A spike is generated when $u$ reaches $V_{th}$ at time $t^f$, and $u$ is reset to the resting potential $u=u_{rest}$, which is usually set to be zero. The output spike train is defined using the Dirac delta function: $s(t) = \sum_{t^f}\delta(t-t^f)$. 

A spiking neural network is composed of connected spiking neurons with connection coefficients. We consider a simple current model $I_i(t)=\sum_j w_{ij}s_j(t) +b_i$, where the subscript $i$ represents the $i$-th neuron, $w_{ij}$ is the weight from neuron $j$ to neuron $i$, and $b_i$ is a bias. The discrete computational form is: 
\vspace{-2mm}
\begin{equation}
    \left\{
    \begin{aligned}
        &u_i\left[t + 1\right] = \lambda (u_i[t] - V_{th}s_i[t]) + \sum_j w_{ij}s_j[t] + b_i,\\
        &s_i[t + 1] = H(u_i\left[t+1\right] - V_{th}),\\
    \end{aligned}
    \right.
    \label{eq.discrete}
\end{equation}
where $H(x)$ is the Heaviside step function,  $s_i[t]$ is the spike train of neuron $i$ at discrete time step $t$, and $\lambda<1$ is a leaky term (typically taken as $1-\frac{1}{\tau_m}$). The constant $R$, $\tau_m$, and time step size are absorbed into the weights and bias. The reset operation is implemented by subtraction. 

\subsection{Previous SNN Training Methods}\label{previous training methods}
\vspace{-2mm}

\textbf{Spike Representation.}\quad The (weighted) firing rate or first spiking time of spiking neurons can be proved to follow ANN-like closed-form transformations~\cite{deng2021optimal,zhou2021temporal,wu2021tandem,xiao2021training,meng2022training}. We focus on the weighted firing rate~\cite{xiao2021training,meng2022training} which has connection with OTTT in this work. Define weighted firing rates and weighted average inputs $\mathbf{a}[t]=\frac{\sum_{\tau=1}^t \lambda^{t-\tau}\mathbf{s}[\tau]}{\sum_{\tau=1}^t \lambda^{t-\tau}}$, $\mathbf{\overline{x}}[t]=\frac{\sum_{\tau=0}^t \lambda^{t-\tau}\mathbf{x}[\tau]}{\sum_{\tau=0}^t \lambda^{t-\tau}}$ in the discrete condition. Given convergent weighted average inputs $\mathbf{\overline{x}}[t]\rightarrow \mathbf{x^*}$, it can be proved that $\mathbf{a}[t]\rightarrow \mathbf{a^*} = \sigma\left(\frac{1}{V_{th}}\mathbf{x^*}\right)$ with bounded random error, where $\sigma$ is a clamp function ($\sigma(x)=\min(\max(0, x), 1)$) in the discrete condition while a ReLU function in the continuous condition. 
For feedforward networks, the closed-form mapping between successive layers is established based on weighted firing rate after time $T$: $\mathbf{a}^{l+1}[T] \approx \sigma\left(\frac{1}{V_{th}}\left(\mathbf{W}^l\mathbf{a}^l[T]+\mathbf{b}^{l+1}\right)\right)$, and gradients are calculated with such spike representation: $\frac{\partial L}{\partial \mathbf{W}^l}=\frac{\partial L}{\partial \mathbf{a}^N[T]}\prod_{i=N-1}^{l+1}\frac{\partial \mathbf{a}^{i+1}[T]}{\partial \mathbf{a}^i[T]}\frac{\partial \mathbf{a}^{l+1}[T]}{\partial \mathbf{W}^l}$. For the recurrent condition, $\mathbf{a}[t]$ will converge to an equilibrium state following an implicit fixed-point equation, e.g. $\mathbf{a^*} = \sigma\left(\frac{1}{V_{th}}\left(\mathbf{W}\mathbf{a^*}+\mathbf{F}\mathbf{x^*}+\mathbf{b}\right)\right)$ for a single-layer network with input connections $F$ and contractive recurrent connections $W$, and gradients can be calculated based on implicit differentiation~\cite{xiao2021training}. Let $\mathbf{a}=f_{\bm{\theta}}(\mathbf{a})$ denote the fixed-point equation ($\bm{\theta}$ are parameters). We have $\frac{\partial L}{\partial \bm{\theta}} = \frac{\partial L}{\partial \mathbf{a}[T]} \left(I-J_{f_{\bm{\theta}}}\vert_{\mathbf{a}[T]}\right)^{-1} \frac{\partial f_{\bm{\theta}}(\mathbf{a}[T])}{\partial \bm{\theta}}$, where $J_{f_{\bm{\theta}}}\vert_{\mathbf{a}[T]}=\frac{\partial f_{\bm{\theta}}(\mathbf{a}[T])}{\partial \mathbf{a}[T]}$ is the Jacobian of $f_{\bm{\theta}}$ at $\mathbf{a}[T]$.

\textbf{BPTT with SG.}\quad BPTT unfolds the iterative update equation in Eq.(\ref{eq.discrete}) and backpropagates along the computational graph as shown in Fig.~\ref{bptt_forward}, \ref{bptt_backward}. The gradients with $T$ time steps are calculated by \footnote{Note that we follow the numerator layout convention for derivatives, i.e. $\nabla_{\bm{\theta}}L=\left(\frac{\partial L}{\partial \bm{\theta}}\right)^\top$ is the gradient with the same dimension of $\bm{\theta}$.\label{footnote1}}:
\begin{equation}
\small
    \frac{\partial L}{\partial \mathbf{W}^l}=\sum_{t=1}^T \frac{\partial L}{\partial \mathbf{s}^{l+1}[t]}{\color{red}\frac{\partial \mathbf{s}^{l+1}[t]}{\partial \mathbf{u}^{l+1}[t]}}\left(\frac{\partial \mathbf{u}^{l+1}[t]}{\partial \mathbf{W}^l}+\sum_{\tau < t}\prod_{i=t-1}^{\tau}\left({\color[rgb]{0,0.69,0.31}\frac{\partial \mathbf{u}^{l+1}[i+1]}{\partial \mathbf{u}^{l+1}[i]}}+{\color{cyan}\frac{\partial \mathbf{u}^{l+1}[i+1]}{\partial \mathbf{s}^{l+1}[i]}}{\color{red}\frac{\partial \mathbf{s}^{l+1}[i]}{\partial \mathbf{u}^{l+1}[i]}}\right)\frac{\partial \mathbf{u}^{l+1}[\tau]}{\partial \mathbf{W}^l}\right),
    \label{bptt gradient}
\end{equation}
where $\mathbf{W}^l$ is the weight from layer $l$ to $l+1$ and $L$ is the loss. The non-differentiable terms $\color{red}\frac{\partial \mathbf{s}^{l}[t]}{\partial \mathbf{u}^{l}[t]}$ will be replaced by surrogate derivatives, e.g. derivatives of  rectangular or sigmoid functions~\cite{wu2018spatio}: $\frac{\partial s}{\partial u}=\frac{1}{a_1}\text{sign}\left(\lvert u - V_{th} \rvert < \frac{a_1}{2}\right)$ or $\frac{\partial s}{\partial u}=\frac{1}{a_2}\frac{e^{(V_{th}-u)/a_2}}{(1+e^{(V_{th}-u)/a_2})^2}$, where $a_1$ and $a_2$ are hyperparameters.

\section{Online Training Through Time for SNNs}\label{sec:method}
\vspace{-2mm}

This section contains four sub-sections. In Section~\ref{derivation of ottt}, we introduce our proposed OTTT by decoupling the temporal dependency of BPTT. Then in Section~\ref{connection spike representation}, we further connect the gradients of OTTT and spike representation-based methods, and prove that OTTT can provide a descent direction for optimization, which is not guaranteed by BPTT with SG. In Section~\ref{connection hebbian}, we discuss the relationship between OTTT and the three-factor Hebbian learning rule. Implementation details are presented in Section~\ref{implementation details}.

\subsection{Derivation of Online Training Through Time}\label{derivation of ottt}
\vspace{-2mm}

\begin{figure}
    \centering
    \subfigure[BPTT Forward]{
    \includegraphics[scale=0.089]{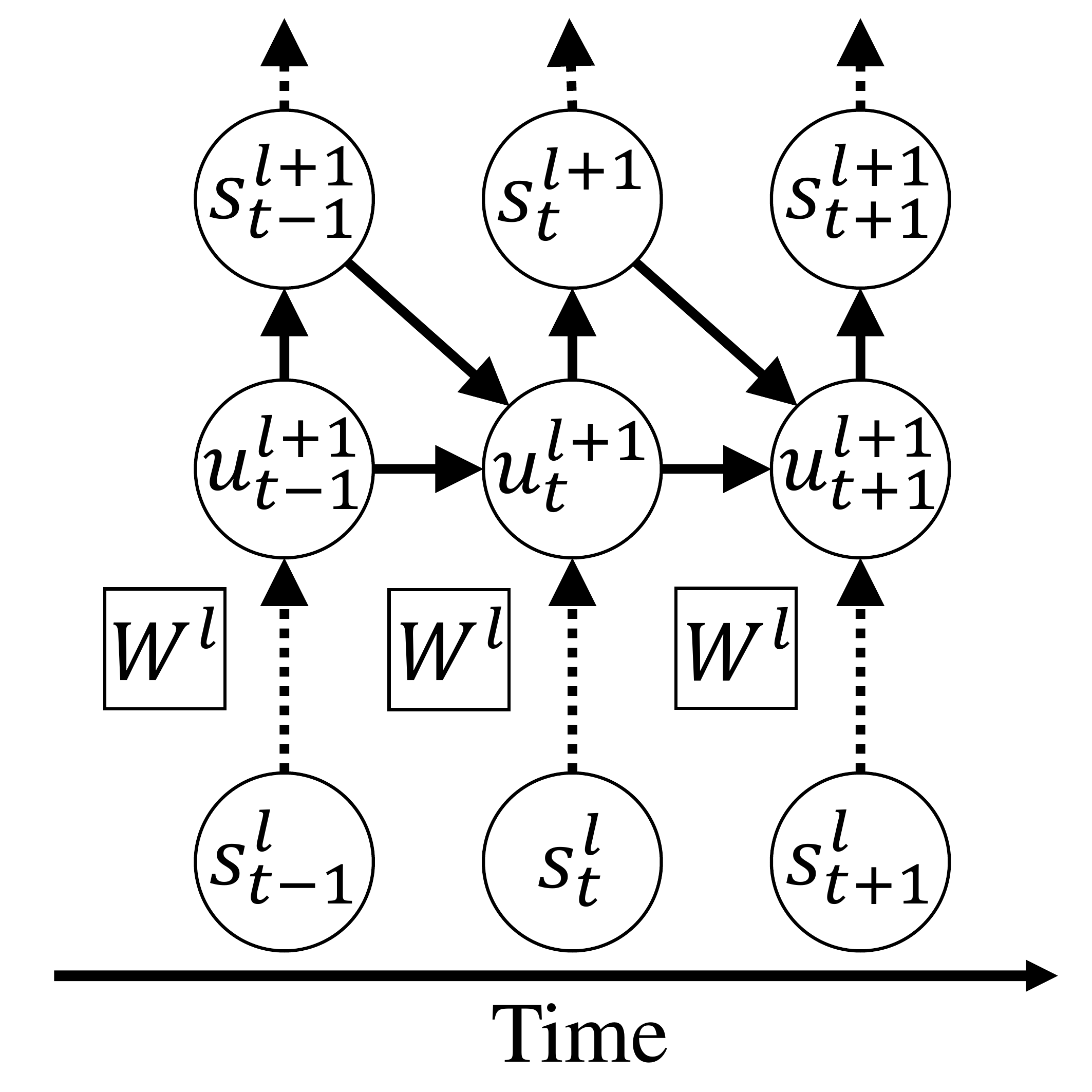}
    \label{bptt_forward}
    }
    \subfigure[OTTT Forward]{
    \includegraphics[scale=0.089]{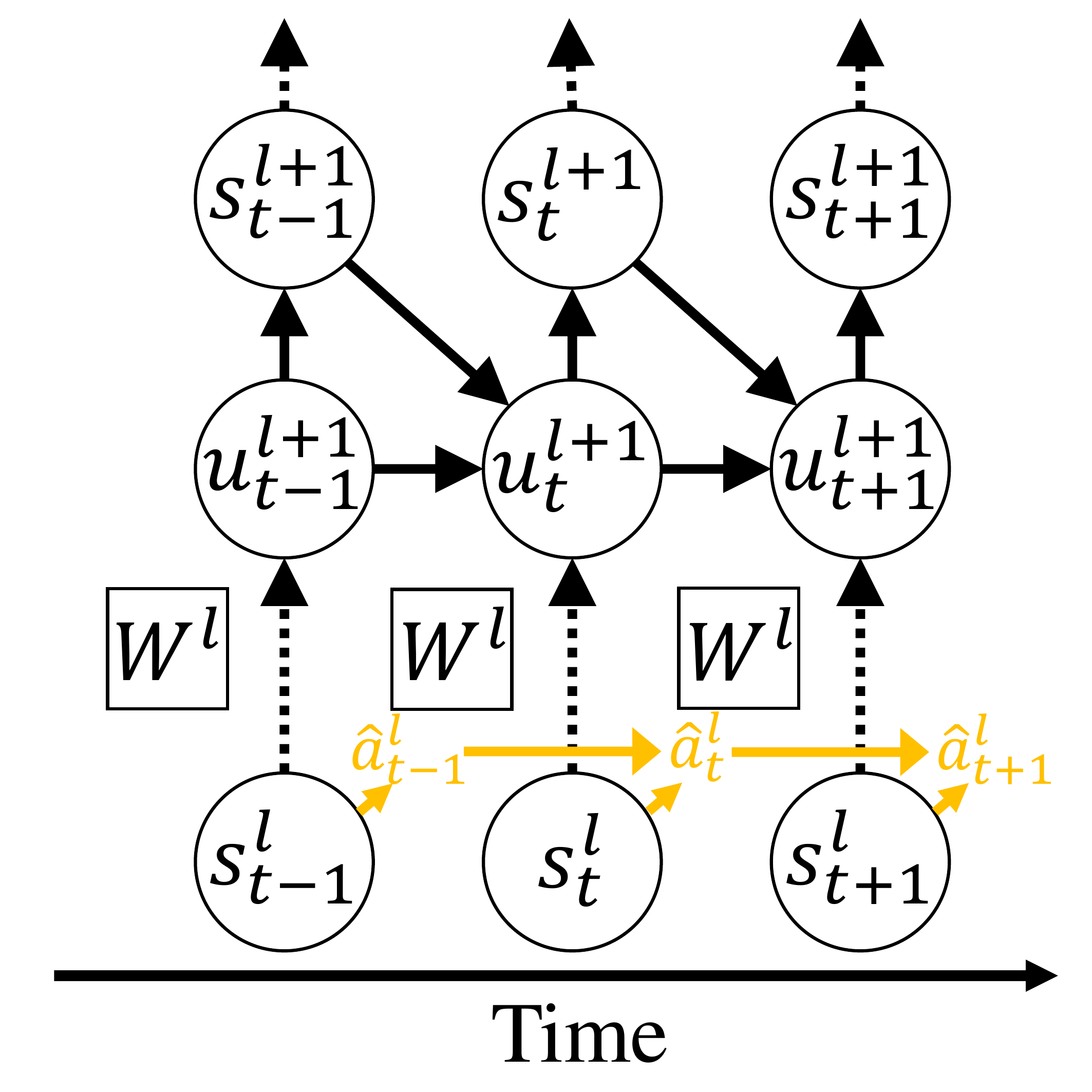}
    \label{ottt_forward}
    }
    \subfigure[BPTT Backward]{
    \includegraphics[scale=0.089]{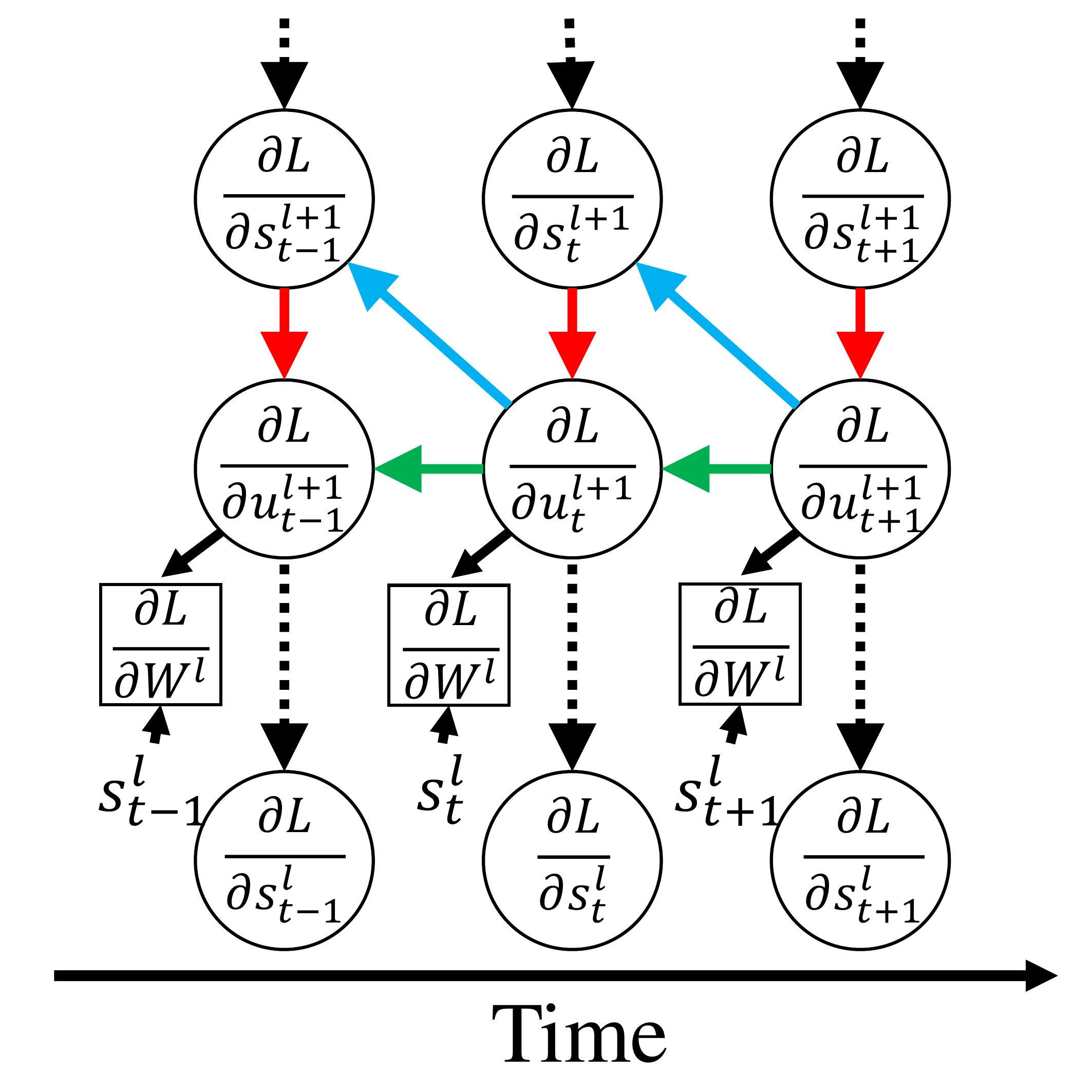}
    \label{bptt_backward}
    }
    \subfigure[OTTT Backward]{
    \includegraphics[scale=0.089]{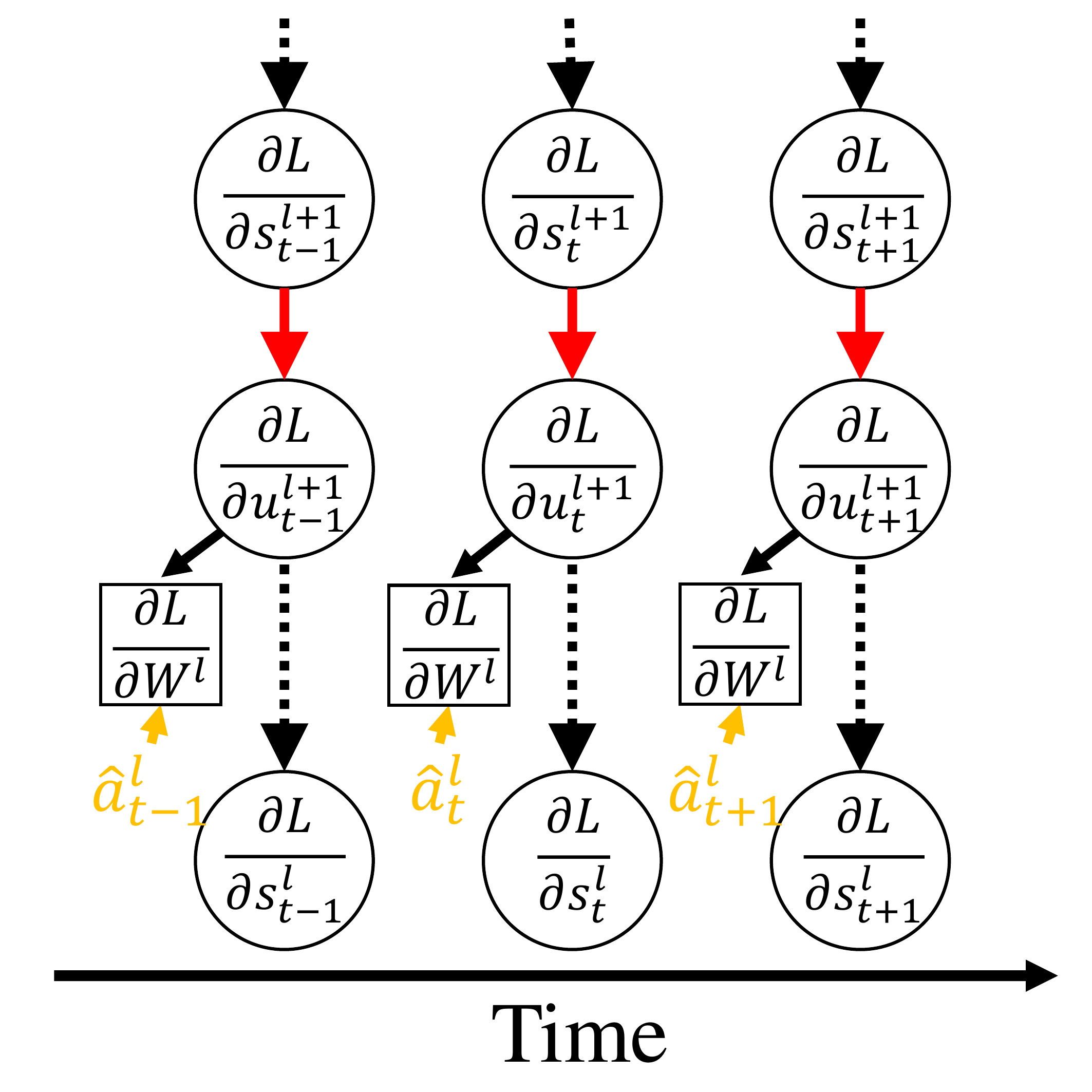}
    \label{ottt_backward}
    }
    \caption{Illustration of the forward and backward procedures of BPTT and OTTT.}
    \label{fig:illustration}
\vspace{-2mm}
\end{figure}

\textbf{Decouple temporal dependency.}\quad As shown in Fig.~\ref{bptt_backward}, BPTT has to maintain the computational graph of previous time to backpropagate through time. We will decouple such temporal dependency to enable online gradient calculation, as illustrated in Fig.~\ref{ottt_backward}. 

We first focus on the feedforward network condition. In this setting, all temporal dependencies lie in the dynamics of each spiking neuron, i.e. {\color[rgb]{0,0.69,0.31}$\frac{\partial \mathbf{u}^{l+1}[i+1]}{\partial \mathbf{u}^{l+1}[i]}$} and ${\color{cyan}\frac{\partial \mathbf{u}^{l+1}[i+1]}{\partial \mathbf{s}^{l+1}[i]}}{\color{red}\frac{\partial \mathbf{s}^{l+1}[i]}{\partial \mathbf{u}^{l+1}[i]}}$ in Eq.(\ref{bptt gradient}). We consider the case that we do not apply surrogate derivatives to ${\color{red}\frac{\partial \mathbf{s}^{l+1}[i]}{\partial \mathbf{u}^{l+1}[i]}}$ in such temporal dependency. Since the derivative of the Heaviside step function is 0 almost everywhere, we have ${\color{cyan}\frac{\partial \mathbf{u}^{l+1}[i+1]}{\partial \mathbf{s}^{l+1}[i]}}{\color{red}\frac{\partial \mathbf{s}^{l+1}[i]}{\partial \mathbf{u}^{l+1}[i]}} \approx 0$\footnote{Note that this is consistent with some released implementations of BPTT with SG methods which detach the neuron reset operation from the computational graph and do not backpropagate gradients in this path~\cite{Fang_2021_ICCV,fang2021deep}.}. Then the dependency only includes {\color[rgb]{0,0.69,0.31}$\frac{\partial \mathbf{u}^{l+1}[i+1]}{\partial \mathbf{u}^{l+1}[i]}$}, which equals $\lambda \mathbf{I}$. 
Therefore, we have\textsuperscript{\ref{footnote1}}:
\begin{equation}
    \frac{\partial L}{\partial \mathbf{W}^l}=\sum_{t=1}^{T} \frac{\partial L}{\partial \mathbf{s}^{l+1}[t]}{\color{red}\frac{\partial \mathbf{s}^{l+1}[t]}{\partial \mathbf{u}^{l+1}[t]}}\left(\sum_{\tau \leq t}\lambda^{t-\tau}\frac{\partial \mathbf{u}^{l+1}[\tau]}{\partial \mathbf{W}^l}\right), \nabla_{\mathbf{W}^l}L = \sum_{t=1}^T \mathbf{g}_{\mathbf{u}^{l+1}}[t]\left(\sum_{\tau \leq t}\lambda^{t-\tau}\mathbf{s}^l[\tau]\right)^\top,
    \label{ottt gradient}
\end{equation}
where $\mathbf{g}_{\mathbf{u}^{l+1}}[t]=\left(\frac{\partial L}{\partial \mathbf{s}^{l+1}[t]}{\color{red}\frac{\partial \mathbf{s}^{l+1}[t]}{\partial \mathbf{u}^{l+1}[t]}}\right)^\top$ is the gradient for $\mathbf{u}^{l+1}[t]$. Based on Eq.(\ref{ottt gradient}), we can track presynaptic activities $\hat{\mathbf{a}}^l[t] = \sum_{\tau \leq t}\lambda^{t-\tau}\mathbf{s}^l[\tau]$ for each neuron during the forward procedure by $\hat{\mathbf{a}}^l[t+1]=\lambda \hat{\mathbf{a}}^l[t]+\mathbf{s}^l[t+1]$, so that when given $\mathbf{g}_{\mathbf{u}^{l+1}}[t]$, gradients at each time step can be calculated independently by $\nabla_{\mathbf{W}^l}L[t]=\mathbf{g}_{\mathbf{u}^{l+1}}[t]{\hat{\mathbf{a}}^l[t]}^\top$ without backpropagation through {\color[rgb]{0,0.69,0.31}$\frac{\partial \mathbf{u}^{l+1}[i+1]}{\partial \mathbf{u}^{l+1}[i]}$}. 

As for the recurrent network condition, there are additional temporal dependencies due to the feedback connections between neurons. If there is feedback connection from layer $l_2$ to $l_1$ ($l_2\geq l_1$), there would be terms such as ${\color{cyan}\frac{\partial \mathbf{u}^{l_1}[i+1]}{\partial \mathbf{s}^{l_2}[i]}}{\color{red}\frac{\partial \mathbf{s}^{l_2}[i]}{\partial \mathbf{u}^{l_2}[i]}}\frac{\partial \mathbf{u}^{l_2}[i]}{\partial \mathbf{u}^{l_1}[i]}$ in the calculation of gradients (note that Eq.~(\ref{bptt gradient}) omit feedback connections for simplicity). We also consider not applying surrogate derivatives to ${\color{red}\frac{\partial \mathbf{s}^{l_2}[i]}{\partial \mathbf{u}^{l_2}[i]}}$ in the temporal dependency so that gradients are not calculated in this path. Similar to the feedforward condition, we can derive that the gradients of the general weight $\mathbf{W}^{l_i\rightarrow l_j}$ from any layer $l_i$ to any layer $l_j$ can be calculated by $\nabla_{\mathbf{W}^{l_i\rightarrow l_j}}L[t]=\mathbf{g}_{\mathbf{u}^{l_j}}[t]{\hat{\mathbf{a}}^{l_i}[t]}^\top$ at each time step. 
A theoretical explanation for optimization will be presented in Section~\ref{connection spike representation}. 

\textbf{Instantaneous Loss and Gradient.}\quad Calculating online gradients, e.g. the above $\mathbf{g}_{\mathbf{u}^{l+1}}[t]$ for $\nabla_{\mathbf{W}^l}L[t]$, requires instantaneous computation of the loss at each time step. Previous typical loss for SNNs is based on the firing rate, e.g. $L_{fr}=\mathcal{L}\left(\frac{1}{T}\sum_{t=1}^T\mathbf{s}^N[t], \mathbf{y}\right)$, where $\mathbf{y}$ is the label, $\mathbf{s}^N[t]$ is the spike at the last layer, and $\mathcal{L}$ can take cross-entropy loss. This loss depends on all time steps and does not support online gradients. We leverage the instantaneous loss and calculate the above $\mathbf{g}_{\mathbf{u}^{l+1}}[t]$ as: 
\vspace{-1mm}
\begin{equation}
    L[t]=\frac{1}{T}\mathcal{L}\left(\mathbf{s}^N[t], \mathbf{y}\right),\quad \mathbf{g}_{\mathbf{u}^{l+1}}[t]=\left(\frac{\partial L[t]}{\partial \mathbf{s}^N[t]}\prod_{i=N-1}^{l+1}\frac{\partial \mathbf{s}^{i+1}[t]}{\partial \mathbf{s}^i[t]} \frac{\partial \mathbf{s}^{l+1}[t]}{\partial \mathbf{u}^{l+1}[t]}\right)^\top.
    \label{instantaneous loss}
\vspace{-2mm}
\end{equation}

The total loss $L\coloneqq\sum_{t=1}^TL[t]$ is the upper bound of $L_{fr}$ when $\mathcal{L}$ is a convex function such as cross-entropy. Then gradients can be calculated independently at each time step, as shown in Fig.~\ref{ottt_backward}. We apply surrogate derivatives for ${\color{red}\frac{\partial \mathbf{s}^{l}[t]}{\partial \mathbf{u}^{l}[t]}}$ in this calculation, which will be explained in Section~\ref{connection spike representation}.

Since gradients are calculated instantaneously at each time step, OTTT does not require maintaining the unfolded computational graph and only requires constant training memory costs agnostic to time steps. Note that instantaneous gradients of OTTT will be different from BPTT with the instantaneous loss for multi-layer or recurrent networks, as we do not consider future influence in the instantaneous calculation: BPTT considers terms such as $\frac{\partial L[t']}{\partial \mathbf{u}^N[t']}\frac{\partial \mathbf{u}^N[t']}{\partial \mathbf{u}^N[t]}\prod_{i=N-1}^l\frac{\partial \mathbf{u}^{i+1}[t]}{\partial \mathbf{u}^i[t]}$ ($t'>t$) for $\mathbf{u}^l[t]$ while we do not. The equivalence of OTTT and BPTT only holds for the last layer, and we do not seek the exact equivalence to BPTT with SG which is theoretically unclear, but will build the connection with spike representations and prove the descent guarantee. Also, note that the tracked presynaptic activities are similar to the biologically plausible ``eligibility traces'' in the literature~\cite{zenke2018superspike, bellec2020solution,murray2019local}, and we will provide a more solid theoretical grounding for optimization in Section~\ref{connection spike representation}.

\subsection{Connection with Spike Representation-Based Methods for Descent Directions}\label{connection spike representation}
\vspace{-2mm}

In this section, we connect gradients of OTTT and spike representation-based methods, and prove that OTTT can provide a descent direction for optimization under feedforward and recurrent conditions with convergent inputs. 

\textbf{Feedforward Networks.}\quad As introduced in Section~\ref{previous training methods}, with convergent inputs, methods based on spike representations establish closed-form mappings between successive layers with weighted firing rate $\mathbf{a}[t]=\frac{\sum_{\tau=1}^t \lambda^{t-\tau}\mathbf{s}[\tau]}{\sum_{\tau=1}^t \lambda^{t-\tau}}$ as $\mathbf{a}^{l+1}[T] \approx \sigma\left(\frac{1}{V_{th}}\left(\mathbf{W}^l\mathbf{a}^l[T]+\mathbf{b}^{l+1}\right)\right)$, and calculate gradients by $\frac{\partial L}{\partial \mathbf{W}^l}=\frac{\partial L}{\partial \mathbf{a}^N[T]}\prod_{i=N-1}^{l+1}\frac{\partial \mathbf{a}^{i+1}[T]}{\partial \mathbf{a}^i[T]}\frac{\partial \mathbf{a}^{l+1}[T]}{\partial \mathbf{W}^l}$. Note that $\mathbf{a}[t]$ is similar to the tracked presynaptic activities $\hat{\mathbf{a}}[t]=\sum_{\tau=1}^t \lambda^{t-\tau}\mathbf{s}[\tau]$ in OTTT. We can obtain: 
\begin{equation}
    \left(\nabla_{\mathbf{W}^l}L_{sr}\right)_{sr} = \sum_{t=1}^T\left(\left(\frac{1}{T}\frac{1}{\lambda^{T-t}}\frac{\partial L_{sr}}{\partial \mathbf{s}^N[t]}\prod_{i=N-1}^{l+1}\frac{\partial \mathbf{a}^{i+1}[T]}{\partial \mathbf{a}^i[T]}\right)^\top\odot \mathbf{d}^{l+1}[T]\right){\hat{\mathbf{a}}^l[T]}^\top,
    \label{sr gradient}
\end{equation}
where $L_{sr}$ is the loss based on spike representation, $\mathbf{d}^{l+1}[T]=\sigma'\left(\frac{1}{V_{th}}\left(\mathbf{W}^l\mathbf{a}^l[T]+\mathbf{b}^{l+1}\right)\right)$, and `$\odot$' is element-wise product. The detailed derivation can be found in Appendix A. 

It can be easily seen that Eq.~(\ref{sr gradient}) has a similar form as gradients in Eq.~(\ref{ottt gradient}), (\ref{instantaneous loss}), and we build the connection between them in three steps. For sake of clarify, in the following, we denote gradients of OTTT and spike representation by $\nabla_{\mathbf{W}^l}L$ and $\left(\nabla_{\mathbf{W}^l}L_{sr}\right)_{sr}$, respectively. 

In the first step, we can use appropriate surrogate derivatives of $\frac{\partial \mathbf{s}^{l+1}_i[t]}{\partial \mathbf{u}^{l+1}_i[t]} (t=1,\cdots,T)$ to approximate $\mathbf{d}^{l+1}_i[T]$. As introduced in Section~\ref{previous training methods}, $\sigma$ is a clamp function ($\sigma(x)=\min(\max(0, x), 1)$) in the discrete condition while a ReLU function in the continuous condition. Then $\mathbf{d}^{l+1}[T]$ almost equals $\text{sign}\left(\lvert \mathbf{W}^l\mathbf{a}^l[T]+\mathbf{b}^{l+1} - V_{th} \rvert < V_{th}\right)$ if we slightly relax the clamp bound caused by the discretization, and this can be approximated by $\text{diag}\left(\frac{\partial \mathbf{s}^{l+1}[t]}{\partial \mathbf{u}^{l+1}[t]}\right)=\text{sign}\left(\lvert \mathbf{u}^{l+1}[t] - V_{th} \rvert < V_{th}\right)$ at each time step. 
Note that surrogate derivatives here approximate the well-defined derivative of the mapping function between $\mathbf{a}[T]$, rather than the pseudo derivative of the non-differentiable Heaviside step function as in BPTT with SG. The approximation is exact except the rare case that averagely a neuron generate spikes (i.e. the average input is positive) while sometimes the membrane potential is less than $u_{rest}$ (or the reverse). For simplicity in the following theoritical analysis, we assume the equivalence between surrogate derivatives and $\mathbf{d}[T]$.
\newtheorem{assumption}{\bf Assumption}
\begin{assumption}\label{assumption1}
$\forall l=1,\cdots,N, t=1,\cdots,T, \rm{diag}\left(\frac{\partial \mathbf{s}^{l+1}[t]}{\partial \mathbf{u}^{l+1}[t]}\right) = \mathbf{d}^{l+1}[T]$.
\end{assumption}

In the second step, we take $L_{sr}$ in Eq.~(\ref{sr gradient}) as $L_{srup}=\frac{1}{\sum_{\tau=0}^{T-1}\lambda^\tau}\sum_{t=1}^T\lambda^{T-t}\mathcal{L}(\mathbf{s}^N[t], \mathbf{y})$ so that $\frac{1}{T}\frac{1}{\lambda^{T-t}}\frac{\partial L_{sr}}{\partial \mathbf{s}^N[t]}$ in Eq.~(\ref{sr gradient}) aligns with $\frac{\partial L[t]}{\partial \mathbf{s}^N[t]}$ in Eq.~(\ref{instantaneous loss}) except a constant term. Note that $L_{srup}$ is an upper bound of the common loss $L_{sr}'=\mathcal{L}(\mathbf{a}^N[T], \mathbf{y})=\mathcal{L}(\frac{\sum_{t=1}^T \lambda^{T-t}\mathbf{s}^N[t]}{\sum_{t=1}^t \lambda^{T-t}}, \mathbf{y})$ if $\mathcal{L}$ is a convex function. 
Then let $\hat{\mathbf{g}}_{\mathbf{u}^{l+1}}[t]=\left(\frac{\partial \mathcal{L}(\mathbf{s}^N[t], \mathbf{y})}{\partial \mathbf{s}^N[t]}\prod_{i=N-1}^{l+1}\frac{\partial \mathbf{s}^{i+1}[t]}{\partial \mathbf{s}^i[t]} \frac{\partial \mathbf{s}^{l+1}[t]}{\partial \mathbf{u}^{l+1}[t]}\right)^\top$, with Assumption~\ref{assumption1} we have: $\nabla_{\mathbf{W}^l}L = \frac{1}{T} \sum_{t=1}^T \hat{\mathbf{g}}_{\mathbf{u}^{l+1}}[t] {\hat{\mathbf{a}}^l[t]}^\top$ and $\left(\nabla_{\mathbf{W}^l}L_{sr}\right)_{sr}=\frac{1}{T}\frac{1}{\sum_{\tau=0}^{T-1}\lambda^\tau}\sum_{t=1}^T\hat{\mathbf{g}}_{\mathbf{u}^{l+1}}[t] {\hat{\mathbf{a}}^l[T]}^\top$. Please refer to Appendix A for details.

In the third step, we handle the remaining difference that $\nabla_{\mathbf{W}^l}L$ leverages instantaneous presynaptic activities $\hat{\mathbf{a}}[t]$ during calculation while $\left(\nabla_{\mathbf{W}^l}L_{sr}\right)_{sr}$ uses the final $\hat{\mathbf{a}}^l[T]$ after time $T$. Note that the weighted firing rate gradually converges $\mathbf{a}[t]\rightarrow \mathbf{a}^*$ with bounded random error. Suppose the errors $\bm{\epsilon}^l[t]=\mathbf{a}^l[t]-\mathbf{a}^l[T]$ are small ($l=0$ represents inputs, i.e. $\mathbf{a}^0[t]=\overline{\mathbf{x}}[t]$), then we have that $-\nabla_{\mathbf{W}^l}L$ can provide a descent direction, as shown in Theorem~\ref{thm_feedforward}. 
\newtheorem{thm}{\bf Theorem}
\begin{thm}\label{thm_feedforward}
If Assumption~\ref{assumption1} holds, $V_{th}=1$, and the errors $\bm{\epsilon}^l[t]=\mathbf{a}^l[t]-\mathbf{a}^l[T]$ are small such that $\left\lVert \sum_{t=1}^T \hat{\mathbf{g}}_{\mathbf{u}^{l+1}}[t] {\bm{\epsilon}^l[t]}^\top \right\rVert < \left\lVert \sum_{t=1}^T \hat{\mathbf{g}}_{\mathbf{u}^{l+1}}[t] {\mathbf{a}^l[T]}^\top \right\rVert - \left\lVert \sum_{t=1}^T \frac{\lambda^t(1-\lambda^{T-t})}{1-\lambda^T}\hat{\mathbf{g}}_{\mathbf{u}^{l+1}}[t] {\mathbf{a}^l[t]}^\top \right\rVert$ when $\left(\nabla_{\mathbf{W}^l}L_{sr}\right)_{sr}\neq\mathbf{0}$, then we have $\left<\nabla_{\mathbf{W}^l}L, \left(\nabla_{\mathbf{W}^l}L_{sr}\right)_{sr}\right> > 0$.
\end{thm}
For the proof and discussion of the assumption please refer to Appendix A. 
With this conclusion, we can explain the descent direction of gradient descent by OTTT for the optimization problem formulated by spike representation. Some random error can be viewed as randomness for stochastic optimization.

\textbf{Recurrent Networks.}\quad For networks with feedback connections, we first consider the single-layer condition for simplicity (see Appendix A for general conditions). We consider feedforward connections $\mathbf{F}$ from inputs to neurons and contractive recurrent connections $\mathbf{W}$ between neurons. As introduced in Section~\ref{previous training methods}, given convergent inputs $\mathbf{\overline{x}}[t]\rightarrow \mathbf{x^*}$, $\mathbf{a}[t]$ of neurons will converge to an equilibrium state $\mathbf{a^*} = f_{\bm{\theta}}(\mathbf{a}^*) =  \sigma\left(\frac{1}{V_{th}}\left(\mathbf{W}\mathbf{a^*}+\mathbf{F}\mathbf{x^*}+\mathbf{b}\right)\right)$ with bounded random error, and gradients are calculated as $\left(\nabla_{\bm{\theta}}L_{sr}\right)_{sr}=\left(\frac{\partial L_{sr}}{\partial \bm{\theta}}\right)^\top = \left(\frac{\partial L_{sr}}{\partial \mathbf{a}[T]} \left(I-J_{f_{\bm{\theta}}}\vert_{\mathbf{a}[T]}\right)^{-1} \frac{\partial f_{\bm{\theta}}(\mathbf{a}[T])}{\partial \bm{\theta}}\right)^\top$, where $\bm{\theta} \in \{\mathbf{W}, \mathbf{F}, \mathbf{b}\}$. 
We consider replacing the inverse Jacobian by an identity matrix: $\widetilde{\left(\nabla_{\bm{\theta}}L_{sr}\right)_{sr}} = \left(\frac{\partial L_{sr}}{\partial \mathbf{a}[T]} \frac{\partial f_{\bm{\theta}}(\mathbf{a}[T])}{\partial \bm{\theta}}\right)^\top$. 
Previous works have proved that this gradient can provide a descent direction for the optimization problem~\cite{fung2021jfb,geng2021training}, i.e. $\left<\widetilde{\left(\nabla_{\bm{\theta}}L_{sr}\right)_{sr}}, \left(\nabla_{\bm{\theta}}L_{sr}\right)_{sr}\right> > 0$. It has a similar form as the OTTT gradient:  $\nabla_{\mathbf{W}}L=\sum_{t=1}^T\mathbf{g}_{\mathbf{u}}[t]{\hat{\mathbf{a}}[t]}^\top$ and $\widetilde{\left(\nabla_{\mathbf{W}}L_{sr}\right)_{sr}}=\sum_{t=1}^T\left(\frac{1}{T}\frac{1}{\lambda^{T-t}}\frac{\partial L_{sr}}{\partial \mathbf{s}[t]}^\top\odot \mathbf{d}[T]\right){\hat{\mathbf{a}}[T]}^\top$. 
Similarly, we can prove the descent guarantee for OTTT as shown in Theorem~\ref{thm_recurrent}. 
For details refer to Appendix A.
\begin{thm}\label{thm_recurrent}
If Assumption~\ref{assumption1} holds, $V_{th}=1$, $\left\lVert J_{f_{\bm{\theta}}}\vert_{\mathbf{a}[T]} \right\rVert \leq \eta < \frac{\sigma_{\text{min}}^2}{\sigma_{\text{max}}^2}$, where $\sigma_{\text{max}}$ and $\sigma_{\text{min}}$ are the maximal and minimal singular value of $\frac{\partial f_{\bm{\theta}}}{\partial \bm{\theta}}\vert_{\mathbf{a}[T]}$, and the errors $\bm{\epsilon}^1[t]=\mathbf{a}[t]-\mathbf{a}[T], \bm{\epsilon}^0[t]=\overline{\mathbf{x}}[t]-\overline{\mathbf{x}}[T]$ are small such that $\left\lVert \sum_{t=1}^T \hat{\mathbf{g}}_{\mathbf{u}}[t] {\bm{\epsilon}^l[t]}^\top  \right\rVert < \frac{\sigma_{\text{min}}^2-\eta \sigma_{\text{max}}^2}{\sigma_{\text{max}}}\left\lVert \sum_{t=1}^T \frac{\partial \mathcal{L}(\mathbf{s}[t], \mathbf{y})}{\partial \mathbf{s}[t]} \left(I-J_{f_{\bm{\theta}}}\vert_{\mathbf{a}[T]}\right)^{-1} \right\rVert - \left\lVert \sum_{t=1}^T \frac{\lambda^t(1-\lambda^{T-t})}{1-\lambda^T}\hat{\mathbf{g}}_{\mathbf{u}}[t] {\mathbf{a}^l[t]}^\top  \right\rVert$ (where $l=0,1$, $\mathbf{a}^1[t]$ and $\mathbf{a}^0[t]$ represent $\mathbf{a}[t]$ and $\overline{\mathbf{x}}[t]$, respectively) when $\left(\nabla_{\bm{\theta}}L_{sr}\right)_{sr}\neq\mathbf{0}$, then we have $\left<\nabla_{\bm{\theta}}L, \left(\nabla_{\bm{\theta}}L_{sr}\right)_{sr}\right> > 0$, where $\bm{\theta}$ are parameters in the network.
\end{thm}
\vspace{-1mm}

\vspace{-2mm}
\subsection{Connection with Three-factor Hebbian Learning Rule}\label{connection hebbian}
\vspace{-2mm}
By explicitly writing the instantaneous gradients of OTTT for the general weight from layer $l_i$ to $l_j$,  $\nabla_{\mathbf{W}^{l_i\rightarrow l_j}}L[t]=\mathbf{g}_{\mathbf{u}^{l_j}}[t]{\hat{\mathbf{a}}^{l_i}[t]}^\top$, and dive into connections between any two neurons $i$ and $j$, we have:
\vspace{-1.5mm}
\begin{equation}
    \nabla_{W_{i,j}}L[t] = \hat{a}_i[t] f(u_j[t]) \delta_j[t],
\end{equation}
where $\hat{a}_i[t]$ is the tracked presynaptic activity, $f(u_j[t])$ is the surrogate derivative function which can represent the change rate of the postsynaptic activity as analyzed in Section~\ref{connection spike representation}, and $\delta_j[t]=g_{s_j}[t]$ is the gradient for neuron output $s_j[t]$ which represents a global modulator. This is a kind of three-factor Hebbian learning rule~\cite{fremaux2016neuromodulated} and the weight can be updated locally with a global signal. The error signal $\delta_j[t]$ can be propagated in an error feedback path simultaneously with feedforward propagation, which is shown biologically plausible with high-frequency bursts~\cite{payeur2021burst}. Note that the analysis in Section~\ref{connection spike representation} still holds if we consider the delay of the propagation of the error signal, i.e. the update is based on $\hat{a}_i[t+\Delta t] f(u_j[t+\Delta t]) \delta_j[t]$.

\vspace{-2mm}
\subsection{Implementation Details}\label{implementation details}
\vspace{-2mm}

As introduced in Section~\ref{derivation of ottt}, we will calculate instantaneous gradients $\nabla_{\mathbf{W}^{l_i\rightarrow l_j}}L[t]=\mathbf{g}_{\mathbf{u}^{l_j}}[t]{\hat{\mathbf{a}}^{l_i}[t]}^\top$ at each time step. We can choose to immediately update parameters before the calculation of the next time step, which we denote as OTTT$_O$, or we can accumulate the gradients by $T$ time steps and then update parameters, which we denote as OTTT$_A$. For OTTT$_O$, we assume that the online update is small and has negligible affects for the following calculation. Pseudo-codes are in Appendix B.

An important issue in practice is that previous BPTT with SG works leverage batch normalization (BN) along the temporal dimension to achieve high performance with extremely low latency on large-scale datasets~\cite{zheng2020going,li2021differentiable,fang2021deep,deng2021temporal}, which requires calculating the mean and variance statistics for all time steps during the forward procedure. This technique intrinsically prevents online gradients and has to suffer from large memory costs. To overcome this shortcoming, we do not use BN, but borrow the idea from normalization-free ResNets (NF-ResNets)~\cite{brock2021characterizing,brock2021high} to replace batch normalization by scaled weight standardization (sWS)~\cite{qiao2019weight}. sWS standardizes weights by $\hat{\mathbf{W}}_{i,j}=\gamma \cdot \frac{\mathbf{W}_{i,j}-\mu_{\mathbf{W}_{i,\cdot}}}{\sigma_{\mathbf{W}_{i,\cdot}}\sqrt{N}}$, and the scale $\gamma$ is determined by analyzing the signal propagation with different activation functions. We apply sWS for VGG~\cite{simonyan2014very} and NF-ResNet architectures in our experiments. For details please refer to Appendix C.

\section{Experiments}\label{sec:exp}
\vspace{-2mm}

In this section, we conduct extensive experiments on CIFAR-10~\cite{krizhevsky2009learning}, CIFAR100~\cite{krizhevsky2009learning}, ImageNet~\cite{deng2009imagenet}, CIFAR10-DVS~\cite{li2017cifar10}, and DVS128-Gesture~\cite{amir2017low} to demonstrate the superior performance of our proposed method on large-scale static and neuromorphic datasets. We leverage the VGG network architecture (64C3-128C3-AP2-256C3-256C3-AP2-512C3-512C3-AP2-512C3-512C3-GAP-FC) for experiments on CIFAR-10, CIFAR-100, CIFAR10-DVS, and DVS128-Gesture, and the NF-ResNet-34~\cite{brock2021characterizing} network architecture for experiments on ImageNet. For all our SNN models, we set $V_{th}=1$ and $\lambda=0.5$. Please refer to Appendix C for training details.

\subsection{Comparison of Training Memory Costs}
\vspace{-2mm}

\begin{figure}[h]
\vspace{-1.5mm}
    \centering
    \includegraphics[scale=0.6]{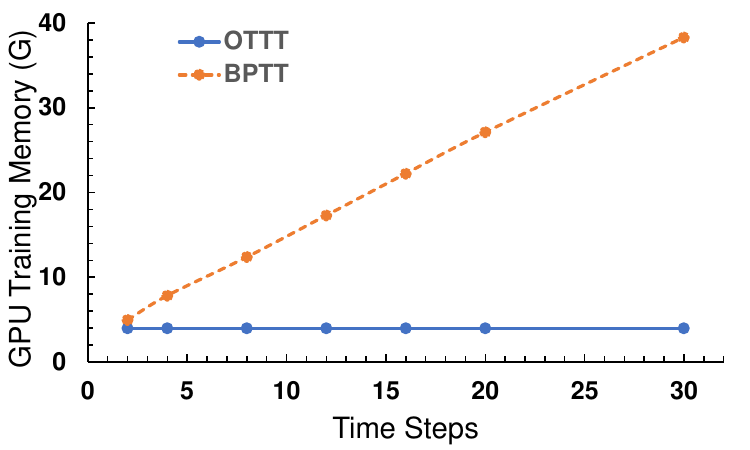}
    \caption{Comparison of training memory costs between OTTT and BPTT under different time steps.}
    \label{fig:training memory}
\vspace{-1.5mm}
\end{figure}

A major advantage of OTTT over BPTT is that OTTT does not require backpropagation along the temporal dimension and therefore only requires constant training memory costs agnostic to time steps, which avoids the large memory costs of BPTT. We verify this by training the VGG network on CIFAR-10 with batch size 128 under different time steps and calculating the memory costs on the GPU. As shown in Fig.~\ref{fig:training memory}, the training memory of BPTT grows linearly with time steps, while OTTT maintains the constant memory (both OTTT$_A$ and OTTT$_O$). Even with a small number of time steps, e.g. 6, OTTT can reduce the memory costs by $2\sim 3\times$. This advantage may also allow training acceleration of SNNs by larger batch sizes with the same computational resources.

\vspace{-2mm}
\begin{table} [ht]
	\centering
	\small
	\tabcolsep=0.5mm
	\caption{Performance on CIFAR-10, CIFAR-100, ImageNet, and CIFAR10-DVS. Results are based on 3 runs of experiments (except ImageNet). Our OTTT is mainly compared with BPTT under the same settings, and is also compared with other representative conversion and direct training methods.}
	\begin{tabular}{c|ccccc}
		\toprule[1pt]
		Dataset & Method & Network structure & Params & Time steps & Mean$\pm$Std (Best) \\
		\midrule[0.5pt]
		\multirow{7}*{CIFAR-10} & ANN-SNN~\cite{deng2021optimal} & VGG-16 & 40M & 16 & (92.29\%)\\
		& BPTT~\cite{zheng2020going} & ResNet-19 (tdBN) & 14.5M & 6 & (93.16\%)\\
		& BPTT~\cite{Fang_2021_ICCV} & 9-layer CNN (PLIF, BN) & 36M & 8 & (93.50\%)\\
		\cline{2-6}
		& BPTT & VGG (sWS) & 9.2M & 6 & 92.78$\pm$0.34\% (93.23\%)\\
		\cline{2-6}
		& \textbf{OTTT$_A$ (ours)} & VGG (sWS) & 9.2M & 6 & \textbf{93.52$\pm$0.06\% (93.58\%)}\\
		& \textbf{OTTT$_O$ (ours)} & VGG (sWS) & 9.2M & 6 & \textbf{93.49$\pm$0.17\% (93.73\%)}\\
		\cline{2-6}
		& ANN & VGG (sWS) & 9.2M & N.A. & (94.43\%)\\
		\midrule[0.5pt]
		\multirow{7}*{CIFAR-100} & ANN-SNN~\cite{deng2021optimal} & VGG-16 & 40M & 400-600 & (70.55\%)\\
		& Hybrid Training~\cite{rathi2019enabling} & VGG-11 & 36M & 125 & (67.87\%)\\
		& DIET-SNN~\cite{rathi2021diet} & VGG-16 & 40M & 5 & (69.67\%)\\
		\cline{2-6}
		& BPTT & VGG (sWS) & 9.3M & 6 & 69.06$\pm$0.07\% (69.15\%)\\
		\cline{2-6}
		& \textbf{OTTT$_A$ (ours)} & VGG (sWS) & 9.3M & 6 & \textbf{71.05$\pm$0.04\% (71.11\%)}\\
		& \textbf{OTTT$_O$ (ours)} & VGG (sWS) & 9.3M & 6 & \textbf{71.05$\pm$0.06\% (71.11\%)}\\
		\cline{2-6}
		& ANN & VGG (sWS) & 9.3M & N.A. & (73.19\%)\\
		\midrule[0.5pt]
		\multirow{5}*{ImageNet} & ANN-SNN~\cite{li2021free} & ResNet-34 & 22M & 32 & (64.54\%)\\
		& Hybrid Training~\cite{rathi2019enabling} & ResNet-34 & 22M & 250 & (61.48\%)\\
		& BPTT~\cite{zheng2020going} & ResNet-34 (tdBN) & 22M & 6 & (63.72\%)\\
		\cline{2-6}
		& \textbf{OTTT$_A$ (ours)} & NF-ResNet-34 & 22M & 6 & \textbf{(65.15\%)}\\
		& \textbf{OTTT$_O$ (ours)} & NF-ResNet-34 & 22M & 6 & \textbf{(64.16\%)}\\
		\midrule[0.5pt]
		\multirow{6}*{DVS-CIFAR10} & Tandem Learning~\cite{wu2021tandem} & CifarNet & 45M & 20 & (65.59\%)\\
		& BPTT~\cite{zheng2020going} & ResNet-19 (tdBN) & 14.5M & 10 & (67.80\%)\\
		& BPTT~\cite{Fang_2021_ICCV} & 7-layer CNN (PLIF, BN) & 1.1M & 20 & (74.80\%)\\
		\cline{2-6}
		& BPTT & VGG (sWS) & 9.2M & 10 & 72.60$\pm$1.26\% (73.90\%)\\
		\cline{2-6}
		& \textbf{OTTT$_A$ (ours)} & VGG (sWS) & 9.2M & 10 & \textbf{76.27$\pm$0.05\% (76.30\%)}\\
		& \textbf{OTTT$_O$ (ours)} & VGG (sWS) & 9.2M & 10 & \textbf{76.63$\pm$0.34\% (77.10\%)}\\
		\bottomrule[1pt]
	\end{tabular}
	\label{performance comparison}
	\vspace{-3mm}
\end{table}

\subsection{Comparison of Performance}
\vspace{-2mm}
We conduct experiments on both large-scale static and neuromorphic datasets. We first verify the effectiveness of the surrogate derivative $\frac{\partial \mathbf{s}^{l+1}[t]}{\partial \mathbf{u}^{l+1}[t]}=\text{sign}\left(\lvert \mathbf{u}^{l+1}[t] - V_{th} \rvert < V_{th}\right)$ in Section~\ref{connection spike representation}. For the VGG (sWS) network on CIFAR-10, OTTT$_A$ and OTTT$_O$ achieves 91.46\% and 92.47\% test accuracy respectively. We empirically observe that applying the sigmoid-like surrogate derivative achieves a higher generalization performance, e.g. OTTT$_A$ and OTTT$_O$ achieves 93.58\% and 93.73\% test accuracy respectively under the same random seed, and a possible reason may be that this introduces some noise for the approximation to regularize the training and improve the generalization. Therefore, in the following performance evaluation, we take sigmoid-like surrogate derivative for OTTT. As shown in Table~\ref{performance comparison}, both OTTT$_A$ and OTTT$_O$ achieve satisfactory performance on all datasets, and compared with BPTT under the same training settings, OTTT achieves higher performance. The proposed OTTT also achieves promising performance on all datasets compared with other representative conversion and direct training methods. Besides, it shows that the performance gap between our SNN model and ANN is around 0.7\% and 2.08\% on CIFAR-10 and CIFAR-100, respectively. Usually, SNNs with a very small number of time steps do not reach the performance of equivalent ANNs due to the information propagation with discrete spikes rather than floating-point numbers. The results of our model with 6 time steps are competitive.

\begin{table}
	\centering
	\small
	\tabcolsep=2mm
	\captionof{table}{Performance on DVS128-Gesture.}
	\begin{tabular}{cccc}
		\toprule[1pt]
		Method & Network structure & Time steps & Accuracy\\
		\midrule[0.5pt]
		SLAYER~\cite{shrestha2018slayer} & 8-layer CNN & 300 & 93.64$\pm$0.49\%\\
		DECOLLE~\cite{kaiser2020synaptic} & 3-layer CNN & 1800 & 95.54$\pm$0.16\% \\
		BPTT~\cite{Fang_2021_ICCV} & 8-layer CNN (PLIF, BN) & 20 & 97.57\% \\
		BPTT~\cite{Fang_2021_ICCV} & 8-layer CNN (LIF, BN) & 20 & 96.88\% \\
		\hline
		BPTT & VGG (sWS) & 20 & 96.88\% \\
		\hline
		\textbf{OTTT$_A$ (ours)} & VGG (sWS) & 20 & 96.88\%\\
		\bottomrule[1pt]
	\end{tabular}
	\label{dvs128 gesture}
    \vspace{-3mm}
\end{table}

We also evaluate our method on the DVS128-Gesture dataset, which is more time-varying with different hand gestures recorded by a DVS camera. As shown in Table~\ref{dvs128 gesture}, our method can achieve the same high performance as BPTT. While our theoretical analysis mainly focus on convergent inputs (e.g. static images or neuromorphic inputs converted from images like CIFAR10-DVS), the results show that our method can also work well for time-varying inputs.

\vspace{-2mm}
\subsection{Effectiveness for Recurrence}
\vspace{-2.5mm}
As introduced in Section~\ref{sec:method}, the proposed OTTT is also valid for networks with feedback connections. Previous works have shown that adding feedback connections can improve the performance of SNNs without much additional costs, especially on the CIFAR-100 datasets~\cite{xiao2021training,kim2022neural}. Therefore, we conduct experiments on CIFAR-100 with the VGG-F network architecture which simply adds a feedback connection from the last feature layer to the first feature layer following~\cite{xiao2021training}, and this weight is zero-intialized. As shown in Table~\ref{feedback network}, the training of VGG-F is valid and VGG-F achieves a higher performance than VGG due to the introduction of feedback connections. Results in Appendix D show that the improvement of OTTT from feedback connections is more significant than that of BPTT. The architectures with feedback connections can be further improved with neural architecture search~\cite{kim2022neural}.

\vspace{-2mm}
\subsection{Effectiveness for Training with Batch Size 1}
\vspace{-2.5mm}

To further study the online training, i.e. not only online in time but also one sample per training, which is consistent with biological learning and learning on neuromorphic hardware, we verify the effectiveness for training with batch size 1. The VGG network on CIFAR-10 is studied, and batch size 1 is compared with the default batch size 128 under the same random seed. Models are only trained for 20 epochs due to the relatively long training time with batch size 1. As shown in Table~\ref{batch size 1}, training with one sample per iteration is still valid, indicating the potential to conduct full online training with the proposed OTTT.

\begin{minipage}{0.48\linewidth}
\newcommand{\tabincell}[2]{\begin{tabular}{@{}#1@{}}#2\end{tabular}}
	\centering
	\small
	\tabcolsep=0.5mm
	\captionof{table}{Performance on CIFAR-100 for VGG and VGG-F trained by OTTT$_O$. Results are based on 3 runs of experiments.}
	\begin{tabular}{ccc}
		\toprule[1pt]
		Network structure & Params & Mean$\pm$Std (Best)\\
		\midrule[0.5pt]
		VGG & 9.3M & 71.05$\pm$0.06\% (71.11\%)\\
		VGG-F & 9.6M & 72.63$\pm$0.23\% (72.94\%)\\
		\bottomrule[1pt]
	\end{tabular}
	\label{feedback network}
\end{minipage}
\hspace{4mm}
\begin{minipage}{0.48\linewidth}
	\centering
	\small
	\tabcolsep=0.5mm
	\captionof{table}{Performance of VGG on CIFAR-10 with different batch sizes for 20 epochs under the same random seed.}
	\begin{tabular}{ccc}
		\toprule[1pt]
		Method & Batch Size & Accuracy\\
		\midrule[0.5pt]
		OTTT$_A$ / OTTT$_O$ & 128 & 88.20\% / 88.62\%\\
		OTTT$_A$ / OTTT$_O$ & 1 & 88.07\% / 88.50\%\\
		\bottomrule[1pt]
	\end{tabular}
	\label{batch size 1}
\end{minipage}

\subsection{Influence of Inference Time Steps}
\vspace{-2mm}
We study the influence of inference time steps on ImageNet as shown in Fig.~\ref{inference time step}. It illustrates that the model trained with time step 6 can achieve higher performance with more inference time steps.

\subsection{Firing Rate Statistics}
\vspace{-2mm}
We study the firing rate statistics of the models trained by OTTT and BPTT, as shown in Fig.~\ref{firing rate}. It demonstrates that models trained by OTTT have higher firing rates in first layers while lower firing rates in later layers compared with BPTT. Overall the firing rate is around 0.19 and with 6 time steps each neuron averagely generate 1.1 spikes, indicating the low energy consumption. Considering that each neuron has more synaptic operations in later layers than first layers (because the channel size is increasing with layers), the synaptic operations of models trained by OTTT and BPTT are about the same ($1.98\times10^8$ vs $1.93\times10^8$). More results please refer to Appendix D.

\begin{minipage}{0.48\linewidth}
	\centering
	\small
	\captionof{figure}{Influence of inference time steps for the model trained with 6 time steps on ImageNet.}
	\includegraphics[width=1\textwidth]{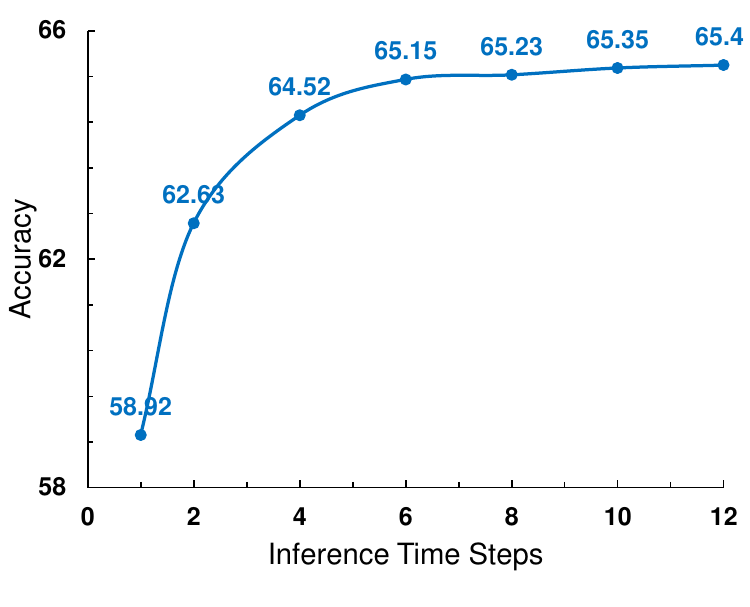}
	\label{inference time step}
\end{minipage}
\hspace{4mm}
\begin{minipage}{0.48\linewidth}
	\centering
	\small
	\captionof{figure}{The average firing rates for the models trained by OTTT and BPTT on CIFAR-10.}
	\includegraphics[width=1\textwidth]{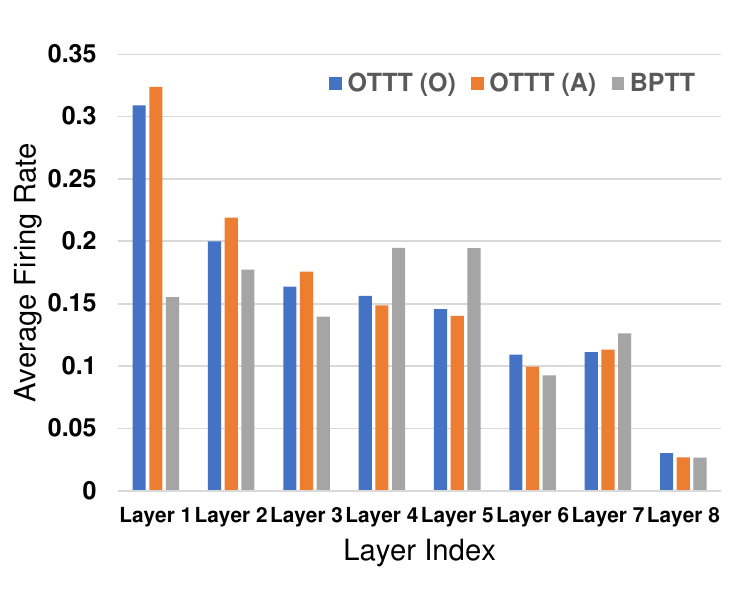}
	\label{firing rate}
\end{minipage}

\section{Conclusion}
\vspace{-2mm}

In this work, we propose a new training method, online training through time (OTTT), for spiking neural networks. We first derive OTTT from BPTT with SG by decoupling the temporal dependency with the tracked pre-synaptic activities, which only requires constant training memory agnostic to time steps and avoids the large training memory costs of BPTT. Then we theoretically analyze and connect the gradients of OTTT and gradients of methods based on spike representations, and prove the descent guarantee of OTTT for the optimization problem under both feedforward and recurrent network conditions. Additionally, we show that OTTT is in the form of three-factor Hebbian learning rule, which is the first to connect BPTT with SG, spike representation-based methods, and biological learning rules. Extensive experiments demonstrate the superior performance of our methods on large-scale static and neuromorphic datasets in a small number of time steps.

\section*{Acknowledgement} 
Z. Lin was supported by the major key project of PCL (No. PCL2021A12), the NSF China (No.s 62276004 and 61731018), and Project 2020BD006 supported by PKU-Baidu Fund.

\small
\bibliographystyle{unsrt}
\bibliography{OTTT}

\section*{Checklist}

\begin{enumerate}

\item For all authors...
\begin{enumerate}
  \item Do the main claims made in the abstract and introduction accurately reflect the paper's contributions and scope?
    \answerYes{}
  \item Did you describe the limitations of your work?
    \answerYes{See Appendix E.}
  \item Did you discuss any potential negative societal impacts of your work?
    \answerYes{See Appendix E.}
  \item Have you read the ethics review guidelines and ensured that your paper conforms to them?
    \answerYes{}
\end{enumerate}

\item If you are including theoretical results...
\begin{enumerate}
  \item Did you state the full set of assumptions of all theoretical results?
    \answerYes{See Section~\ref{sec:method}.}
	\item Did you include complete proofs of all theoretical results?
    \answerYes{See Appendix A.}
\end{enumerate}

\item If you ran experiments...
\begin{enumerate}
  \item Did you include the code, data, and instructions needed to reproduce the main experimental results (either in the supplemental material or as a URL)?
    \answerYes{See Supplementary Materials.}
  \item Did you specify all the training details (e.g., data splits, hyperparameters, how they were chosen)?
    \answerYes{See Appendix C.}
	\item Did you report error bars (e.g., with respect to the random seed after running experiments multiple times)?
    \answerYes{See Section~\ref{sec:exp}.}
	\item Did you include the total amount of compute and the type of resources used (e.g., type of GPUs, internal cluster, or cloud provider)?
    \answerYes{See Appendix C.}
\end{enumerate}

\item If you are using existing assets (e.g., code, data, models) or curating/releasing new assets...
\begin{enumerate}
  \item If your work uses existing assets, did you cite the creators?
    \answerYes{See Section~\ref{sec:exp}.}
  \item Did you mention the license of the assets?
    \answerYes{See Appendix C.}
  \item Did you include any new assets either in the supplemental material or as a URL?
    \answerNA{We do not use new assets.}
  \item Did you discuss whether and how consent was obtained from people whose data you're using/curating?
    \answerNo{We use the existing common datasets.}
  \item Did you discuss whether the data you are using/curating contains personally identifiable information or offensive content?
    \answerNo{We use the existing common datasets.}
\end{enumerate}

\item If you used crowdsourcing or conducted research with human subjects...
\begin{enumerate}
  \item Did you include the full text of instructions given to participants and screenshots, if applicable?
    \answerNA{}
  \item Did you describe any potential participant risks, with links to Institutional Review Board (IRB) approvals, if applicable?
    \answerNA{}
  \item Did you include the estimated hourly wage paid to participants and the total amount spent on participant compensation?
    \answerNA{}
\end{enumerate}

\end{enumerate}


\newpage

\appendix

\section{Detailed Derivation and Proofs}

\subsection{Derivation of Eq. (6)}

Since $\mathbf{a}[t]=\frac{\sum_{\tau=1}^t \lambda^{t-\tau}\mathbf{s}[\tau]}{\sum_{\tau=1}^t \lambda^{t-\tau}}, \hat{\mathbf{a}}[t]=\sum_{\tau=1}^t \lambda^{t-\tau}\mathbf{s}[\tau], \mathbf{a}^{l+1}[T] \approx \sigma\left(\frac{1}{V_{th}}\left(\mathbf{W}^l\mathbf{a}^l[T]+\mathbf{b}^{l+1}\right)\right)$, $\mathbf{d}^{l+1}[T]=\sigma'\left(\frac{1}{V_{th}}\left(\mathbf{W}^l\mathbf{a}^l[T]+\mathbf{b}^{l+1}\right)\right)$, $\left(\frac{\partial L_{sr}}{\partial \mathbf{W}^l}\right)_{sr}=\frac{\partial L_{sr}}{\partial \mathbf{a}^N[T]}\prod_{i=N-1}^{l+1}\frac{\partial \mathbf{a}^{i+1}[T]}{\partial \mathbf{a}^i[T]}\frac{\partial \mathbf{a}^{l+1}[T]}{\partial \mathbf{W}^l}$, and we have $\frac{\partial L_{sr}}{\partial \hat{\mathbf{a}}^N[T]}=\frac{1}{\lambda^{T-t}}\frac{\partial L_{sr}}{\partial \mathbf{s}^N[t]} (\forall 1\leq t\leq T)$\footnote{Note that we can treat $\mathbf{s}^N[t]$ independent with each other, if we consider taking the derivative of the Heaviside step function as 0 in this calculation and therefore $\frac{\partial \mathbf{s}^N[t+1]}{\partial \mathbf{s}^N[t]}=\frac{\partial \mathbf{s}^N[t+1]}{\partial \mathbf{u}^N[t+1]}\frac{\partial \mathbf{u}^N[t+1]}{\partial \mathbf{s}^N[t]}=0$.}, $\frac{\partial L_{sr}}{\partial \hat{\mathbf{a}}^N[T]}=\frac{1}{T}\sum_{t=1}^T\frac{1}{\lambda^{T-t}}\frac{\partial L_{sr}}{\partial \mathbf{s}^N[t]}$, we can obtain:
\begin{equation}
\begin{aligned}
    \left(\nabla_{\mathbf{W}^l}L_{sr}\right)_{sr} &= \left(\frac{\partial L_{sr}}{\partial \mathbf{W}^l}\right)_{sr}^\top = \left(\frac{\partial L_{sr}}{\partial \mathbf{a}^N[T]}\prod_{i=N-1}^{l+1}\frac{\partial \mathbf{a}^{i+1}[T]}{\partial \mathbf{a}^i[T]}\frac{\partial \mathbf{a}^{l+1}[T]}{\partial \mathbf{W}^l}\right)^\top\\
    &= \left(\left(\frac{\partial L_{sr}}{\partial \mathbf{a}^N[T]}\prod_{i=N-1}^{l+1}\frac{\partial \mathbf{a}^{i+1}[T]}{\partial \mathbf{a}^i[T]}\right)^\top\odot \mathbf{d}^{l+1}[T]\right){\mathbf{a}^l[T]}^\top\\
    &= \left(\left(\frac{\partial L_{sr}}{\partial \hat{\mathbf{a}}^N[T]}\prod_{i=N-1}^{l+1}\frac{\partial \mathbf{a}^{i+1}[T]}{\partial \mathbf{a}^i[T]}\right)^\top\odot \mathbf{d}^{l+1}[T]\right){\hat{\mathbf{a}}^l[T]}^\top\\
    &= \sum_{t=1}^T\left(\left(\frac{1}{T}\frac{1}{\lambda^{T-t}}\frac{\partial L_{sr}}{\partial \mathbf{s}^N[t]}\prod_{i=N-1}^{l+1}\frac{\partial \mathbf{a}^{i+1}[T]}{\partial \mathbf{a}^i[T]}\right)^\top\odot \mathbf{d}^{l+1}[T]\right){\hat{\mathbf{a}}^l[T]}^\top.
\end{aligned}
\end{equation}

\subsection{A few notes for the notation of time for multi-layer networks}

Please note that the notation of discrete time steps for multi-layer networks may be slightly different from Eq. (2). We use $s^{i+1}[t]$ to denote the $(i+1)$-th layer’s response after receiving the $i$-th layer’s signals $s^i[t]$. Rigorously speaking, there will be synaptic delay $t_d$ for information propagation between two layers if we consider the whole network in an asynchronous way, so the precise time of $s^{i+1}[t]$ or $u^{i+1}[t]$ for layer $i+1$ may be $t+t_d$ compared with $s^i[t]$ for layer $i$. To simplify the notations, we use $0, 1, \cdots T$ for each layer to represent the corresponding discrete time steps, while the actual time of different layers at time step $t$ should consider some delay across layers.

\subsection{Proof of Theorem 1}

In this subsection, we prove Theorem~\ref{supthm_feedforward} with Assumption~\ref{supassumption1}.

\newtheorem{supassumption}{\bf Assumption}
\begin{supassumption}\label{supassumption1}
$\forall l=1,\cdots,N, t=1,\cdots,T, \rm{diag}\left(\frac{\partial \mathbf{s}^{l+1}[t]}{\partial \mathbf{u}^{l+1}[t]}\right) = \mathbf{d}^{l+1}[T]$.
\end{supassumption}

\newtheorem{supthm}{\bf Theorem}
\begin{supthm}\label{supthm_feedforward}
If Assumption~\ref{supassumption1} holds, $V_{th}=1$, and the errors $\bm{\epsilon}^l[t]=\mathbf{a}^l[t]-\mathbf{a}^l[T]$ are small such that $\left\lVert \sum_{t=1}^T \hat{\mathbf{g}}_{\mathbf{u}^{l+1}}[t] {\bm{\epsilon}^l[t]}^\top \right\rVert < \left\lVert \sum_{t=1}^T \hat{\mathbf{g}}_{\mathbf{u}^{l+1}}[t] {\mathbf{a}^l[T]}^\top \right\rVert - \left\lVert \sum_{t=1}^T \frac{\lambda^t(1-\lambda^{T-t})}{1-\lambda^T}\hat{\mathbf{g}}_{\mathbf{u}^{l+1}}[t] {\mathbf{a}^l[t]}^\top \right\rVert$ when $\left(\nabla_{\mathbf{W}^l}L_{sr}\right)_{sr}\neq\mathbf{0}$, then we have $\left<\nabla_{\mathbf{W}^l}L, \left(\nabla_{\mathbf{W}^l}L_{sr}\right)_{sr}\right> > 0$.
\end{supthm}

\begin{proof}

As described in Sections 4.1 and 4.2, for gradients of OTTT, we have $\nabla_{\mathbf{W}^l}L=\sum_{t=1}^T\mathbf{g}_{\mathbf{u}^{l+1}}[t]{\hat{\mathbf{a}}^l[t]}^\top$, $L\coloneqq\sum_{t=1}^TL[t]=\sum_{t=1}^T\frac{1}{T}\mathcal{L}\left(\mathbf{s}^N[t], \mathbf{y}\right),\quad \mathbf{g}_{\mathbf{u}^{l+1}}[t]=\left(\frac{\partial L[t]}{\partial \mathbf{s}^N[t]}\prod_{i=N-1}^{l+1}\frac{\partial \mathbf{s}^{i+1}[t]}{\partial \mathbf{s}^i[t]} \frac{\partial \mathbf{s}^{l+1}[t]}{\partial \mathbf{u}^{l+1}[t]}\right)^\top$; for gradients based on spike representation, we have $\left(\nabla_{\mathbf{W}^l}L_{sr}\right)_{sr} = \sum_{t=1}^T\left(\left(\frac{1}{T}\frac{1}{\lambda^{T-t}}\frac{\partial L_{sr}}{\partial \mathbf{s}^N[t]}\prod_{i=N-1}^{l+1}\frac{\partial \mathbf{a}^{i+1}[T]}{\partial \mathbf{a}^i[T]}\right)^\top\odot \mathbf{d}^{l+1}[T]\right){\hat{\mathbf{a}}^l[T]}^\top$ and consider $L_{sr}=\frac{1}{\sum_{\tau=0}^{T-1}\lambda^\tau}\sum_{t=1}^T\lambda^{T-t}\mathcal{L}(\mathbf{s}^N[t], \mathbf{y})$. Let $\hat{\mathbf{g}}_{\mathbf{u}^{l+1}}[t]=\left(\frac{\partial \mathcal{L}(\mathbf{s}^N[t], \mathbf{y})}{\partial \mathbf{s}^N[t]}\prod_{i=N-1}^{l+1}\frac{\partial \mathbf{s}^{i+1}[t]}{\partial \mathbf{s}^i[t]} \frac{\partial \mathbf{s}^{l+1}[t]}{\partial \mathbf{u}^{l+1}[t]}\right)^\top$, we have $\nabla_{\mathbf{W}^l}L = \frac{1}{T} \sum_{t=1}^T \hat{\mathbf{g}}_{\mathbf{u}^{l+1}}[t] {\hat{\mathbf{a}}^l[t]}^\top$. With Assumption~\ref{supassumption1}, we have $\frac{\partial \mathbf{s}^{l+1}_i[t]}{\partial \mathbf{u}^{l+1}_i[t]}=\mathbf{d}^{l+1}_i[T]$ and thus $\frac{\partial \mathbf{s}^{l+1}[t]}{\partial \mathbf{s}^{l}[t]}=\frac{\partial \mathbf{a}^{l+1}[t]}{\partial \mathbf{a}^{l}[t]}$ (because $\frac{\partial \mathbf{s}^{l+1}_j[t]}{\partial \mathbf{s}^{l}_i[t]}=\frac{\partial \mathbf{s}^{l+1}_j[t]}{\partial \mathbf{u}^{l+1}_j[t]}\cdot\mathbf{W}_{i,j}=\mathbf{d}^{l+1}_j[T]\cdot\mathbf{W}_{i,j}=\frac{\partial \mathbf{a}^{l+1}_j[t]}{\partial \mathbf{a}^{l}_i[t]}$). So we can derive that $\left(\nabla_{\mathbf{W}^l}L_{sr}\right)_{sr}=\frac{1}{T}\frac{1}{\sum_{\tau=0}^{T-1}\lambda^\tau}\sum_{t=1}^T\hat{\mathbf{g}}_{\mathbf{u}^{l+1}}[t] {\hat{\mathbf{a}}^l[T]}^\top=\frac{1}{T}\sum_{t=1}^T\hat{\mathbf{g}}_{\mathbf{u}^{l+1}}[t] {\mathbf{a}^l[T]}^\top$.

We consider $\widehat{\nabla_{\mathbf{W}^l}L} = \frac{1}{\sum_{\tau=0}^{T-1}\lambda^\tau}\nabla_{\mathbf{W}^l}L = \frac{1}{T}\frac{1}{\sum_{\tau=0}^{T-1}\lambda^\tau} \sum_{t=1}^T \hat{\mathbf{g}}_{\mathbf{u}^{l+1}}[t] {\hat{\mathbf{a}}^l[t]}^\top=\frac{1}{T}\sum_{t=1}^T \hat{\mathbf{g}}_{\mathbf{u}^{l+1}}[t] \frac{\sum_{\tau=0}^{t-1}\lambda^\tau}{\sum_{\tau=0}^{T-1}\lambda^\tau}{\mathbf{a}^l[t]}^\top$. Since the errors $\bm{\epsilon}^l[t]=\mathbf{a}^l[t]-\mathbf{a}^l[T]$ are small such that $\left\lVert \sum_{t=1}^T \hat{\mathbf{g}}_{\mathbf{u}^{l+1}}[t] {\bm{\epsilon}^l[t]}^\top \right\rVert < \left\lVert \sum_{t=1}^T \hat{\mathbf{g}}_{\mathbf{u}^{l+1}}[t] {\mathbf{a}^l[T]}^\top \right\rVert - \left\lVert \sum_{t=1}^T \frac{\lambda^t(1-\lambda^{T-t})}{1-\lambda^T}\hat{\mathbf{g}}_{\mathbf{u}^{l+1}}[t] {\mathbf{a}^l[t]}^\top \right\rVert$ when $\left(\nabla_{\mathbf{W}^l}L_{sr}\right)_{sr}\neq\mathbf{0}$, we have:
\begin{equation}
\begin{aligned}
    \left\lVert \widehat{\nabla_{\mathbf{W}^l}L}-\left(\nabla_{\mathbf{W}^l}L_{sr}\right)_{sr} \right\rVert &= \left\lVert \frac{1}{T}\sum_{t=1}^T \hat{\mathbf{g}}_{\mathbf{u}^{l+1}}[t] \left(\frac{\sum_{\tau=0}^{t-1}\lambda^\tau}{\sum_{\tau=0}^{T-1}\lambda^\tau}{\mathbf{a}^l[t]}^\top - {\mathbf{a}^l[T]}^\top\right) \right\rVert\\
    &= \left\lVert \frac{1}{T}\sum_{t=1}^T \hat{\mathbf{g}}_{\mathbf{u}^{l+1}}[t] \left(\bm{\epsilon}^l[t] - \frac{\lambda^t(1-\lambda^{T-t})}{1-\lambda^T}{\mathbf{a}^l[t]}^\top\right) \right\rVert\\
    &\leq \left\lVert \frac{1}{T}\sum_{t=1}^T \hat{\mathbf{g}}_{\mathbf{u}^{l+1}}[t] \bm{\epsilon}^l[t] \right\rVert + \left\lVert \frac{1}{T}\sum_{t=1}^T \hat{\mathbf{g}}_{\mathbf{u}^{l+1}}[t]\frac{\lambda^t(1-\lambda^{T-t})}{1-\lambda^T}{\mathbf{a}^l[t]}^\top \right\rVert\\
    &< \left\lVert \frac{1}{T}\sum_{t=1}^T \hat{\mathbf{g}}_{\mathbf{u}^{l+1}}[t] {\mathbf{a}^l[T]}^\top \right\rVert = \left\lVert \left(\nabla_{\mathbf{W}^l}L_{sr}\right)_{sr} \right\rVert.
\end{aligned}
\end{equation}

Then, we can obtain:
\begin{equation}
\begin{aligned}
    \left<\widehat{\nabla_{\mathbf{W}^l}L}, \left(\nabla_{\mathbf{W}^l}L_{sr}\right)_{sr}\right> &= \left<\widehat{\nabla_{\mathbf{W}^l}L}-\left(\nabla_{\mathbf{W}^l}L_{sr}\right)_{sr}, \left(\nabla_{\mathbf{W}^l}L_{sr}\right)_{sr}\right> + \left\lVert \left(\nabla_{\mathbf{W}^l}L_{sr}\right)_{sr} \right\rVert^2\\
    & \geq \left\lVert \left(\nabla_{\mathbf{W}^l}L_{sr}\right)_{sr} \right\rVert^2 - \left\lVert \widehat{\nabla_{\mathbf{W}^l}L}-\left(\nabla_{\mathbf{W}^l}L_{sr}\right)_{sr} \right\rVert \left\lVert \left(\nabla_{\mathbf{W}^l}L_{sr}\right)_{sr} \right\rVert > 0.
\end{aligned}
\end{equation}

Therefore, $\left<\nabla_{\mathbf{W}^l}L, \left(\nabla_{\mathbf{W}^l}L_{sr}\right)_{sr}\right> = \left(\sum_{\tau=0}^{T-1}\lambda^\tau\right)\left<\widehat{\nabla_{\mathbf{W}^l}L}, \left(\nabla_{\mathbf{W}^l}L_{sr}\right)_{sr}\right> > 0$.

\end{proof}

\newtheorem{remark}{\bf Remark}
\begin{remark}
As for the assumption of the errors in the theorem, since the weighted firing rate gradually converges $\mathbf{a}[t]\rightarrow \mathbf{a}^*$ with bounded random error caused by the remaining membrane potential at the last time step, the order of errors $\bm{\epsilon}^l[t]$ would be smaller than $\mathbf{a}^l[T]$ especially when $t$ is large. And $\frac{\lambda^t(1-\lambda^{T-t})}{1-\lambda^T} \rightarrow 0$ with $t\rightarrow T$ is also a small number on the right side of the inequality. So this is a reasonable assumption.
\end{remark}

\begin{remark}
The above conclusion mainly focuses on the gradients for connection weights $\mathbf{W}^l$. As for other parameters such as biases $\mathbf{b}^l$, the gradients of OTTT do not involve pre-synaptic activities, so under Assumption~\ref{supassumption1} they are exactly the same as gradients based on spike representation except a constant scaling factor $\frac{1}{\sum_{\tau=0}^{T-1}\lambda^\tau}$.
\end{remark}

\begin{remark}
Note that the gradients based on spike representation may also include small errors since the calculation of SNN is not exactly the same as the equivalent ANN-like mappings. And a larger time step may lead to more accurate gradients. We connect the gradients of OTTT and gradients based on spike representation to demonstrate the overall descent direction, and it is tolerant to small errors, which can also be viewed as randomness for stochastic optimization.
\end{remark}

\subsection{Proof of Theorem 2}

In this subsection, we prove Theorem~\ref{supthm_recurrent}. 

\begin{supthm}\label{supthm_recurrent}
If Assumption~\ref{supassumption1} holds, $V_{th}=1$, $\left\lVert J_{f_{\bm{\theta}}}\vert_{\mathbf{a}[T]} \right\rVert \leq \eta < \frac{\sigma_{\text{min}}^2}{\sigma_{\text{max}}^2}$, where $\sigma_{\text{max}}$ and $\sigma_{\text{min}}$ are the maximal and minimal singular value of $\frac{\partial f_{\bm{\theta}}}{\partial \bm{\theta}}\vert_{\mathbf{a}[T]}$, and the errors $\bm{\epsilon}^1[t]=\mathbf{a}[t]-\mathbf{a}[T], \bm{\epsilon}^0[t]=\overline{\mathbf{x}}[t]-\overline{\mathbf{x}}[T]$ are small such that $\left\lVert \sum_{t=1}^T \hat{\mathbf{g}}_{\mathbf{u}}[t] {\bm{\epsilon}^l[t]}^\top  \right\rVert < \frac{\sigma_{\text{min}}^2-\eta \sigma_{\text{max}}^2}{\sigma_{\text{max}}}\left\lVert \sum_{t=1}^T \frac{\partial \mathcal{L}(\mathbf{s}[t], \mathbf{y})}{\partial \mathbf{s}[t]} \left(I-J_{f_{\bm{\theta}}}\vert_{\mathbf{a}[T]}\right)^{-1} \right\rVert - \left\lVert \sum_{t=1}^T \frac{\lambda^t(1-\lambda^{T-t})}{1-\lambda^T}\hat{\mathbf{g}}_{\mathbf{u}}[t] {\mathbf{a}^l[t]}^\top  \right\rVert$ (where $l=0,1$, $\mathbf{a}^1[t]$ and $\mathbf{a}^0[t]$ represent $\mathbf{a}[t]$ and $\overline{\mathbf{x}}[t]$, respectively) when $\left(\nabla_{\bm{\theta}}L_{sr}\right)_{sr}\neq\mathbf{0}$, then we have $\left<\nabla_{\bm{\theta}}L, \left(\nabla_{\bm{\theta}}L_{sr}\right)_{sr}\right> > 0$, where $\bm{\theta}$ are parameters in the network.
\end{supthm}

\begin{proof}

As described in Sections 4.1 and 4.2 and similar to the proof of Theorem 1, let $\hat{\mathbf{g}}_{\mathbf{u}}[t]=\left(\frac{\partial \mathcal{L}(\mathbf{s}[t], \mathbf{y})}{\partial \mathbf{s}[t]}\frac{\partial \mathbf{s}[t]}{\partial \mathbf{u}[t]}\right)^\top$, we have $\nabla_{\mathbf{W}}L = \frac{1}{T} \sum_{t=1}^T \hat{\mathbf{g}}_{\mathbf{u}}[t] {\hat{\mathbf{a}}[t]}^\top$, $\nabla_{\mathbf{F}}L = \frac{1}{T} \sum_{t=1}^T \hat{\mathbf{g}}_{\mathbf{u}}[t] {\hat{\mathbf{x}}[t]}^\top$ (where $\hat{\mathbf{x}}[t]=\sum_{\tau=1}^t \lambda^{t-\tau}\mathbf{x}[\tau]$), and $\nabla_{\mathbf{b}}L = \frac{1}{T} \sum_{t=1}^T \hat{\mathbf{g}}_{\mathbf{u}}[t]$. For gradients based on spike representation, $\left(\nabla_{\bm{\theta}}L_{sr}\right)_{sr}=\left(\frac{\partial L_{sr}}{\partial \mathbf{a}[T]} \left(I-J_{f_{\bm{\theta}}}\vert_{\mathbf{a}[T]}\right)^{-1} \frac{\partial f_{\bm{\theta}}(\mathbf{a}[T])}{\partial \bm{\theta}}\right)^\top$, and we will also consider $\widetilde{\left(\nabla_{\bm{\theta}}L_{sr}\right)_{sr}} = \left(\frac{\partial L_{sr}}{\partial \mathbf{a}[T]} \frac{\partial f_{\bm{\theta}}(\mathbf{a}[T])}{\partial \bm{\theta}}\right)^\top$. Considering $L_{sr}=\frac{1}{\sum_{\tau=0}^{T-1}\lambda^\tau}\sum_{t=1}^T\lambda^{T-t}\mathcal{L}(\mathbf{s}[t], \mathbf{y})$, 
and with Assumption~\ref{supassumption1} which indicates $\frac{\partial \mathbf{s}^{l+1}_i[t]}{\partial \mathbf{u}^{l+1}_i[t]}=\mathbf{d}^{l+1}_i[T]$, we can derive that $\widetilde{\left(\nabla_{\mathbf{W}}L_{sr}\right)_{sr}} = \frac{1}{T} \sum_{t=1}^T \hat{\mathbf{g}}_{\mathbf{u}}[t] {\mathbf{a}}[T]^\top$, $\widetilde{\left(\nabla_{\mathbf{F}}L_{sr}\right)_{sr}} = \frac{1}{T} \sum_{t=1}^T \hat{\mathbf{g}}_{\mathbf{u}}[t] {\overline{\mathbf{x}}}[T]^\top$, and $\widetilde{\left(\nabla_{\mathbf{b}}L_{sr}\right)_{sr}} = \frac{1}{T}\frac{1}{\sum_{\tau=0}^{T-1}\lambda^\tau} \sum_{t=1}^T \hat{\mathbf{g}}_{\mathbf{u}}[t]$.

We consider $\widehat{\nabla_{\mathbf{W}}L} = \frac{1}{\sum_{\tau=0}^{T-1}\lambda^\tau}\nabla_{\mathbf{W}}L = \frac{1}{T}\sum_{t=1}^T \hat{\mathbf{g}}_{\mathbf{u}^{l+1}}[t] \frac{\sum_{\tau=0}^{t-1}\lambda^\tau}{\sum_{\tau=0}^{T-1}\lambda^\tau}{\mathbf{a}^l[t]}^\top$ and $\widehat{\nabla_{\mathbf{F}}L} = \frac{1}{\sum_{\tau=0}^{T-1}\lambda^\tau}\nabla_{\mathbf{F}}L$. Since $\left\lVert J_{f_{\bm{\theta}}}\vert_{\mathbf{a}[T]} \right\rVert \leq \eta < \frac{\sigma_{\text{min}}^2}{\sigma_{\text{max}}^2}$, where $\sigma_{\text{max}}$ and $\sigma_{\text{min}}$ are the maximal and minimal singular value of $\frac{\partial f_{\bm{\theta}}}{\partial \bm{\theta}}\vert_{\mathbf{a}[T]}$ ($\bm{\theta}\in \{\mathbf{W}, \mathbf{F}, \mathbf{b}\}$), and the errors $\bm{\epsilon}[t]=\mathbf{a}[t]-\mathbf{a}[T]$ are small such that $\left\lVert \sum_{t=1}^T \hat{\mathbf{g}}_{\mathbf{u}}[t] {\bm{\epsilon}^l[t]}^\top  \right\rVert < \frac{\sigma_{\text{min}}^2-\eta \sigma_{\text{max}}^2}{\sigma_{\text{max}}}\left\lVert \sum_{t=1}^T \frac{\partial \mathcal{L}(\mathbf{s}[t], \mathbf{y})}{\partial \mathbf{s}[t]} \left(I-J_{f_{\bm{\theta}}}\vert_{\mathbf{a}[T]}\right)^{-1} \right\rVert - \left\lVert \sum_{t=1}^T \frac{\lambda^t(1-\lambda^{T-t})}{1-\lambda^T}\hat{\mathbf{g}}_{\mathbf{u}}[t] {\mathbf{a}[t]}^\top  \right\rVert$ when $\left(\nabla_{\bm{\theta}}L\right)_{sr}\neq\mathbf{0}$, we can obtain: 
\begin{equation}
\begin{aligned}
    \left\lVert \widehat{\nabla_{\mathbf{W}}L} - \widetilde{\left(\nabla_{\mathbf{W}}L_{sr}\right)_{sr}} \right\rVert &= \left\lVert \frac{1}{T}\sum_{t=1}^T \hat{\mathbf{g}}_{\mathbf{u}}[t] \left(\frac{\sum_{\tau=0}^{t-1}\lambda^\tau}{\sum_{\tau=0}^{T-1}\lambda^\tau}{\mathbf{a}[t]}^\top - {\mathbf{a}[T]}^\top\right) \right\rVert\\
    &= \left\lVert \frac{1}{T}\sum_{t=1}^T \hat{\mathbf{g}}_{\mathbf{u}}[t] \left(\bm{\epsilon}[t] - \frac{\lambda^t(1-\lambda^{T-t})}{1-\lambda^T}{\mathbf{a}[t]}^\top\right) \right\rVert\\
    &\leq \left\lVert \frac{1}{T}\sum_{t=1}^T \hat{\mathbf{g}}_{\mathbf{u}}[t] \bm{\epsilon}[t] \right\rVert + \left\lVert \frac{1}{T}\sum_{t=1}^T \hat{\mathbf{g}}_{\mathbf{u}}[t]\frac{\lambda^t(1-\lambda^{T-t})}{1-\lambda^T}{\mathbf{a}[t]}^\top \right\rVert\\
    &< \frac{\sigma_{\text{min}}^2-\eta \sigma_{\text{max}}^2}{\sigma_{\text{max}}}\left\lVert \frac{1}{T}\sum_{t=1}^T \frac{\partial \mathcal{L}(\mathbf{s}[t], \mathbf{y})}{\partial \mathbf{s}[t]} \left(I-J_{f_{\bm{\theta}}}\vert_{\mathbf{a}[T]}\right)^{-1} \right\rVert.
\end{aligned}
\end{equation}

Then, we have (let $\mathbf{v} = \left(\frac{\partial L_{sr}}{\partial \mathbf{a}[T]} \left(I-J_{f_{\bm{\theta}}}\vert_{\mathbf{a}[T]}\right)^{-1} \right)^\top = \frac{1}{T}\sum_{t=1}^T\left(\frac{\partial \mathcal{L}(\mathbf{s}[t], \mathbf{y})}{\partial \mathbf{s}[t]} \left(I-J_{f_{\bm{\theta}}}\vert_{\mathbf{a}[T]}\right)^{-1}\right)^\top$): 
\begin{align}
    \left<\widehat{\nabla_{\mathbf{W}}L}, \left(\nabla_{\mathbf{W}}L_{sr}\right)_{sr}\right> &= \left<\widetilde{\left(\nabla_{\mathbf{W}}L_{sr}\right)_{sr}}, \left(\nabla_{\mathbf{W}}L_{sr}\right)_{sr}\right> + \left<\widehat{\nabla_{\mathbf{W}}L} - \widetilde{\left(\nabla_{\mathbf{W}}L_{sr}\right)_{sr}}, \left(\nabla_{\mathbf{W}}L_{sr}\right)_{sr}\right>\notag\\
    &= \mathbf{v}^\top \frac{\partial f_{\bm{\theta}}(\mathbf{a}[T])}{\partial \mathbf{W}} \left(\frac{\partial L_{sr}}{\partial \mathbf{a}[T]} \frac{\partial f_{\bm{\theta}}(\mathbf{a}[T])}{\partial \mathbf{W}}\right)^\top + \left<\widehat{\nabla_{\mathbf{W}}L} - \widetilde{\left(\nabla_{\mathbf{W}}L_{sr}\right)_{sr}}, \left(\nabla_{\mathbf{W}}L_{sr}\right)_{sr}\right>\notag\\
    &= \mathbf{v}^\top \frac{\partial f_{\bm{\theta}}(\mathbf{a}[T])}{\partial \mathbf{W}} {\frac{\partial f_{\bm{\theta}}(\mathbf{a}[T])}{\partial \mathbf{W}}}^\top \left(I-J_{f_{\bm{\theta}}}\vert_{\mathbf{a}[T]}\right)^\top \mathbf{v} + \left<\widehat{\nabla_{\mathbf{W}}L} - \widetilde{\left(\nabla_{\mathbf{W}}L_{sr}\right)_{sr}}, \left(\nabla_{\mathbf{W}}L_{sr}\right)_{sr}\right>\notag\\
    &= \left\lVert \mathbf{v}^\top \frac{\partial f_{\bm{\theta}}(\mathbf{a}[T])}{\partial \mathbf{W}} \right\rVert^2 - \mathbf{v}^\top \frac{\partial f_{\bm{\theta}}(\mathbf{a}[T])}{\partial \mathbf{W}} {\frac{\partial f_{\bm{\theta}}(\mathbf{a}[T])}{\partial \mathbf{W}}}^\top {J_{f_{\bm{\theta}}}\vert_{\mathbf{a}[T]}}^\top \mathbf{v}\notag\\
    &+ \left<\widehat{\nabla_{\mathbf{W}}L} - \widetilde{\left(\nabla_{\mathbf{W}}L_{sr}\right)_{sr}}, \left(\nabla_{\mathbf{W}}L_{sr}\right)_{sr}\right>\notag\\
    &\geq \sigma_{\text{min}}^2 \lVert \mathbf{v} \rVert^2 - \eta \sigma_{\text{max}}^2 \lVert \mathbf{v} \rVert^2 - \left\lVert \widehat{\nabla_{\mathbf{W}}L} - \widetilde{\left(\nabla_{\mathbf{W}}L_{sr}\right)_{sr}} \right\rVert  \left\lVert \mathbf{v}^\top \frac{\partial f_{\bm{\theta}}(\mathbf{a}[T])}{\partial \mathbf{W}} \right\rVert \notag\\
    &> \sigma_{\text{min}}^2 \lVert \mathbf{v} \rVert^2 - \eta \sigma_{\text{max}}^2 \lVert \mathbf{v} \rVert^2 - \frac{\sigma_{\text{min}}^2-\eta \sigma_{\text{max}}^2}{\sigma_{\text{max}}}\left\lVert \mathbf{v} \right\rVert \cdot \sigma_{\text{max}} \left\lVert \mathbf{v} \right\rVert = 0.
\end{align}

Therefore, $\left<\nabla_{\mathbf{W}}L, \left(\nabla_{\mathbf{W}}L_{sr}\right)_{sr}\right> = \left(\sum_{\tau=0}^{T-1}\lambda^\tau\right)\left<\widehat{\nabla_{\mathbf{W}}L}, \left(\nabla_{\mathbf{W}}L_{sr}\right)_{sr}\right> > 0$. Similarly, we can derive that $\left<\nabla_{\mathbf{F}}L, \left(\nabla_{\mathbf{F}}L_{sr}\right)_{sr}\right> > 0$. And for $\nabla_{\mathbf{b}}L$, we have $\nabla_{\mathbf{b}}L=\left(\sum_{\tau=0}^{T-1}\lambda^\tau\right)\left(\nabla_{\mathbf{b}}L_{sr}\right)_{sr}$, so $\left<\nabla_{\mathbf{b}}L, \left(\nabla_{\mathbf{b}}L_{sr}\right)_{sr}\right> > 0$. Therefore, for all parameters $\bm{\theta}$ in the network, we have $\left<\nabla_{\bm{\theta}}L, \left(\nabla_{\bm{\theta}}L_{sr}\right)_{sr}\right> > 0$ when $\left(\nabla_{\bm{\theta}}L_{sr}\right)_{sr}\neq\mathbf{0}$.

\end{proof}

\begin{remark}
The above conclusion considers the single-layer condition. It can be generalized to the multi-layer condition. For example, if we consider multiple feedforward hidden layers (denote the weight as $F^l$) with a feedback connection from the last hidden layer to the first hidden layer (denote the weight as $W^1$), and assume the function is contractive, the equilibrium states for each layer are ${\mathbf{a}^1}^* = f_1\left(f_N\circ\cdots\circ f_2({\mathbf{a}^1}^*), \mathbf{x^*}\right)$ and ${\mathbf{a}^{l+1}}^*=f_{l+1}({\mathbf{a}^l}^*)$, where $f_1(\mathbf{a}, \mathbf{x})=\sigma\left(\frac{1}{V_{th}}(\mathbf{W}^1\mathbf{a}+\mathbf{F}^1\mathbf{x}+\mathbf{b}^1)\right)$ and $f_{l}(\mathbf{a}) = \sigma\left(\frac{1}{V_{th}}(\mathbf{F}^{l}\mathbf{a}+\mathbf{b}^{l})\right)$~\cite{xiao2021training}. Then with a similar condition for the Jacobian of $f_{\bm{\theta}}=f_N\circ\cdots\circ f_2\circ f_1$ and errors $\bm{\epsilon}^l[t]$ of each layer as in Theorem~\ref{supthm_recurrent}, we can prove $\left<\nabla_{\bm{\theta}}L, \left(\nabla_{\bm{\theta}}L\right)_{sr}\right> > 0$ when $\left(\nabla_{\bm{\theta}}L\right)_{sr}\neq\mathbf{0}$ for all parameters $\bm{\theta}$ in the network as well.
More generally, multi-layer networks with arbitrary feedback connections can be written in a single-layer formulation, i.e. we consider all neurons in different layers as a whole single layer, and feedforward or feedback connections can be viewed as connections between these neurons, which is written as a much larger weight matrix with some imposed structures representing the connection restrictions. Therefore, the conclusion can be directly generalized to these conditions as well.
\end{remark}

\begin{remark}
The assumption $\left\lVert J_{f_{\bm{\theta}}}\vert_{\mathbf{a}[T]} \right\rVert \leq \eta < \frac{\sigma_{\text{min}}^2}{\sigma_{\text{max}}^2}$ is also made in previous works~\cite{fung2021jfb,geng2021training} and we consider it as a reasonable assumption in the theoretical analysis. It is a sufficient condition to bound the worst case, and in practice it is unnecessary to always enforce the restriction, as indicated in \cite{fung2021jfb}.
\end{remark}

\section{Pseudocode of the OTTT algorithm}

We present the pseudocode of one iteration of OTTT training for a feedforward network in Algorithm~\ref{algorithm: ottt} to better illustrate our training method.

\begin{algorithm}[h]
    \caption{One iteration of OTTT training for a feedforward network.}
    \hspace*{0.02in} {\bf Input:}
    Network parameters $\{\mathbf{W}^l\}, \{\mathbf{b}^l\}$; Input data $x$; Label $y$; Time steps $T$; Other hyperparameters;\\
    \hspace*{0.02in} {\bf Output:}
    Trained network parameters $\{\mathbf{W}^l\}, \{\mathbf{b}^l\}$.\\
    \begin{algorithmic}[1]
    \For{$t=1,2,\cdots, T$}
        \For{$l=1,2,\cdots,N$} \quad // {\bf Forward}
            \State Update membrane potentials $\mathbf{u}^l[t]$ and generate spikes $\mathbf{s}^l[t]$ at layer $l$;
            \State Update the tracked presynaptic activities $\hat{\mathbf{a}}^l[t]=\lambda \hat{\mathbf{a}}^l[t-1]+\hat{\mathbf{s}}^l[t]$ at layer $l$.
        \EndFor
        \For{$l=N,N-1,\cdots,1$} \quad // {\bf Backward}
            \State Calculate the instantaneous backpropagated errors $\mathbf{g}_{\mathbf{u}^l}[t]$;
            \State Calculate the instantaneous gradient $\nabla_{\mathbf{W}^{l-1}}L[t] = \mathbf{g}_{\mathbf{u}^l}[t](\hat{\mathbf{a}}^{l-1}[t])^\top$.
        \If{online update} \quad // OTTT$_O$
            \State Update $\mathbf{W}^{l-1}$ with $\nabla_{\mathbf{W}^{l-1}}L[t]$ based on the gradient-based optimizer;
            \State Update $\mathbf{b}^{l}$ with $\mathbf{g}_{\mathbf{u}^l}[t]$ based on the gradient-based optimizer.
        \Else \quad // OTTT$_A$
            \State Accumulate gradients $\nabla_{\mathbf{W}^{l-1}}L=\nabla_{\mathbf{W}^{l-1}}L+\nabla_{\mathbf{W}^{l-1}}L[t]$, $\nabla_{\mathbf{b}^{l}}L=\nabla_{\mathbf{b}^{l}}L+\mathbf{g}_{\mathbf{u}^l}[t]$.
        \EndIf
        \EndFor
    \EndFor
    \If{not online update} \quad // OTTT$_A$
        \State Update parameters $\{\mathbf{W}^l\}$ with accumulated $\{\nabla_{\mathbf{W}^l}L\}$ based on the gradient-based optimizer;
        \State Update parameters $\{\mathbf{b}^l\}$ with accumulated $\{\nabla_{\mathbf{b}^l}L\}$ based on the gradient-based optimizer.
    \EndIf
    \end{algorithmic}
    \label{algorithm: ottt}
\end{algorithm}

\section{Implementation Details}

\subsection{Scaled Weight Standardization and NF-ResNets}

The scaled weight standardization (sWS) is proposed in \cite{brock2021characterizing,brock2021high} to replace the commonly used batch normalization (BN) and realize normalization-free ResNets (NF-ResNets). Different from BN which standardizes the activation with different samples, sWS standardizes weights by:
\begin{equation}
    \hat{\mathbf{W}}_{i,j}=\gamma \cdot \frac{\mathbf{W}_{i,j}-\mu_{\mathbf{W}_{i,\cdot}}}{\sigma_{\mathbf{W}_{i,\cdot}}\sqrt{N}},
\end{equation}
where $\mu_{\mathbf{W}_{i,\cdot}}$ and $\sigma_{\mathbf{W}_{i,\cdot}}$ are the mean and variance calculated along the input dimension, and the scale $\gamma$ is determined by analyzing the signal propagation with different activation functions. The original weight standardization is proposed in \cite{qiao2019weight}, which is shown to share the similar benefit as BN to smooth the loss landscape, if combined with other normalization techniques, e.g. group normalization. sWS further takes the signal propagation into account so that the variance of the signal is preserved during the forward propagation of neural networks and the mean of the output is 0, which is another property of BN. Particularly, for the input $\mathbf{x}$ that is sampled i.i.d from $\mathcal{N}(0, 1)$, considering the ReLU activation $g$, \cite{brock2021characterizing} derive that we should take $\gamma=\frac{\sqrt{2}}{\sqrt{1-\frac{1}{\pi}}}$ to preserve the variance of signals, i.e. $\text{Var}(\hat{\mathbf{W}}g(\mathbf{x}))=1$. This is because the outputs $g(x)=\max(x, 0)$ with Gaussian inputs will be sampled from the rectified Gaussian distribution with variance $\sigma_g^2=(1/2)(1 - (1/\pi))$~\cite{brock2021characterizing}. In this work, to ensure the variance preserving at each time step of the SNN computation, we derive $\gamma$ based on the consideration of the signals after the Heaviside step function $H$. Particularly, consider the Gaussian input $\mathbf{x}$, when $V_{th}=1$, the variance of the outputs $H(x-V_{th})$ is $\sigma_H^2=\frac{1}{2}\text{erfc}(\frac{1}{\sqrt{2}})\left(1 - \frac{1}{2}\text{erfc}(\frac{1}{\sqrt{2}})\right)$. So we will take $\gamma = \frac{1}{\sigma_H}\approx 2.74$ to preserve the variance of signals. Additionally, \cite{brock2021characterizing} demonstrates that sWS can incorporate another learnable scaling factor for the weights, which is also taken in common BN implementations. Therefore, we also adopt this sWS technique, which is the same as the pseudocode in~\cite{brock2021characterizing}. For VGG network structures, we directly impose sWS on all weights. For NF-ResNet structures, we use the same structure as in~\cite{brock2021characterizing}, which is briefly introduced below.

NF-ResNets~\cite{brock2021characterizing} consider the residual networks $x_{l+1}=x_l+\alpha f_l(x_l/\beta_l)$, which differs from ResNets~\cite{he2016identity} in three aspects: 1. NF-ResNets remove the BN components in ResNets and impose sWS on all weights; 2. a scaling factor $\alpha$ is added for each residual branch; 3. for the input of each residual branch, it will first be divided by the term $\beta_l$ that represents the standard deviation of signals. Note that the third point is because the residual computation $x_{l+1}=x_l+\alpha f_l(x_l/\beta_l)$ will gradually accumulate the variance of the residual branch, i.e. $\text{Var}(x_{l+1})=\text{Var}(x_l)+\text{Var}(\alpha f_l(x_l/\beta_l))$, so dividing $\beta_l$ ensures that the residual branch keeps the identity variance 1 (combined with sWS), and this also indicates to calculate $\beta_l$ by $\beta_{l+1}^2=\beta_l^2+\alpha^2$ after each branch. Also, note that for each transition block, the identity path with a strided conv will also be first divided by $\beta_l$, so the variance is reset after each transition block between two stages. For the implementation details, we mainly follow the pseudocode in~\cite{brock2021characterizing} and replace the activation functions by functions of spiking neurons, and we take $\alpha=0.2$. For more illustrations and other details, please directly refer to~\cite{brock2021characterizing}.

\subsection{Training Settings}

\subsubsection{Datasets}

We conduct experiments on CIFAR-10~\cite{krizhevsky2009learning}, CIFAR-100~\cite{krizhevsky2009learning}, ImageNet~\cite{deng2009imagenet},  CIFAR10-DVS~\cite{li2017cifar10}, and DVS128-Gesture~\cite{amir2017low}.

\paragraph{CIFAR-10} CIFAR-10 is a dataset of color images with 10 classes of objects, which contains 50,000 training samples and 10,000 testing samples. Each sample is a $32\times32\times3$ color image. We normalize the inputs based on the global mean and standard deviation, and apply random cropping, horizontal flipping and cutout~\cite{devries2017improved} for data augmentation. The inputs to the first layer of SNNs at each time step are directly the pixel values, which can be viewed as a real-valued input current.

\paragraph{CIFAR-100} CIFAR-100 is a dataset similar to CIFAR-10 except that there are 100 classes of objects. It also consists of 50,000 training samples and 10,000 testing samples. We use the same pre-processing as CIFAR-10.

The license of CIFAR-10 and CIFAR-100 is the MIT License. 

\paragraph{ImageNet} ImageNet-1K is a dataset of color images with 1000 classes of objects, which contains 1,281,167 training samples and 50,000 validation images. We adopt the common pre-possessing strategies, i.e. the training images are first randomly resized and cropped to $224\times224$, and then normalized after the random horizontal flipping data augmentation, while the testing images are first resized to $256\times256$ and center-cropped to $224\times224$, and then normalized. The inputs are also converted to a real-valued input current at each time step. The license of ImageNet is Custom (non-commercial).

\paragraph{DVS-CIFAR10}
The DVS-CIFAR10 dataset is the neuromorphic version of the CIFAR-10 dataset converted by a Dynamic Vision Sensor (DVS), which is composed of 10,000 samples, one-sixth of the original CIFAR-10. It consists of spike trains with two channels corresponding to ON- and OFF-event spikes. The pixel dimension is expanded to $128\times128$. Following the common practice, we split the dataset into 9000 training samples and 1000 testing samples. As for the data pre-processing, we reduce the time resolution by accumulating the spike events~\cite{Fang_2021_ICCV} into 10 time steps, and we reduce the spatial resolution into $48\times48$ by interpolation. We apply the random cropping augmentation as CIFAR-10 to the input data, and normalize the inputs based on the global mean and standard deviation of all time steps (which can be integrated into the connection weights of the first layer). The license of DVS-CIFAR10 is CC BY 4.0.

\paragraph{DVS128-Gesture}
The DVS128-Gesture dataset is a neuromorphic dataset that contains 11 kinds of hand gestures from 29 subjects under 3 kinds of illumination
conditions recorded by a DVS camera. It is composed of 1176 training samples and 288 testing samples. Following~\cite{Fang_2021_ICCV}, we pre-possess the data to integrate event data into 20 frames. 
The license of DVS128-Gesture is the Creative Commons Attribution 4.0 license.

\subsubsection{Training Hyperparameters}

For our SNN models, we assume the neurons of the last classification layer will not spike or reset, and do classification based on the accumulated membrane potential, which is the same as~\cite{xiao2021training}. That is, the final output is $\mathbf{u}^N[t]=\mathbf{W}^{N-1}\mathbf{s}^{N-1}[t]+\mathbf{b}^N$ at each time step. The classification is based on the accumulated $\mathbf{u}^N=\sum_{t=1}^T\mathbf{u}^N[t]$, and the loss during training is also calculated based on $\mathbf{u}^N[t]$, i.e. $\mathcal{L}(\mathbf{u}^N[t], \mathbf{y})$.

For CIFAR-10, CIFAR-100, and DVS-CIFAR10, models are trained by SGD with momentum 0.9 for 300 epochs with the default batch size 128, and the initial learning rate is set as 0.1 with a cosine annealing learning rate scheduler to 0 (for the experiments of training with batch size 1, the initial learning rate is linearly rescaled to $\frac{0.1}{128}$). For DVS-CIFAR10, we apply dropout on all layers with dropout rate as 0.1. As for the loss function, inspired by \cite{deng2021temporal}, we combine cross-entropy (CE) loss and mean-square-error (MSE) loss, i.e. $\mathcal{L}(\mathbf{u}^N[t], \mathbf{y})=(1-\alpha)\textbf{CE}(\mathbf{u}^N[t], \mathbf{y})+\alpha\textbf{MSE}(\mathbf{u}^N[t], \mathbf{y})$, where $\alpha$ is taken as 0.05 for CIFAR10 and CIFAR100 while 0.001 for DVS-CIFAR10.

For ImageNet, models are trained by SGD with momentum 0.9 for 100 epochs with the default batch size 256, and the initial learning rate is set as 0.1, which is decayed by 0.1 every 30 epochs. We set the weight decay as $2\times 10^{-5}$, and no dropout is applied. The loss function takes the cross-entropy loss.

For DVS128-Gesture, models are trained by the Adam optimizer for 300 epochs with batch size 16, and the initial learning rate is set as 0.001 with a cosine annealing learning rate scheduler to 0. No dropout is applied. As for the loss function, we set $\alpha=0.001$ following DVS-CIFAR10.

The code implementation is based on the PyTorch framework~\cite{paszke2019pytorch}, and experiments are carried out on one NVIDIA GeForce RTX 3090 GPU.

\section{Additional Experiment Results}

\subsection{Firing Rate Statistics on ImageNet}

\begin{figure}[h]
	\centering
	\includegraphics[width=0.8\textwidth]{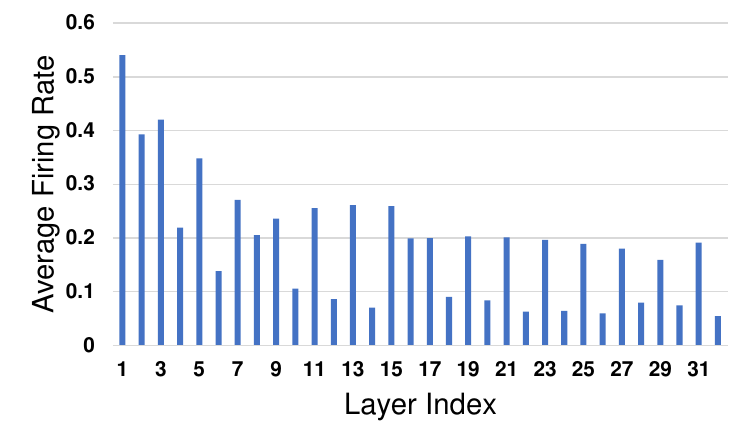}
	\caption{The average firing rates for the model trained by OTTT$_A$ on ImageNet.}
	\label{firing rate imagenet}
\end{figure}

In this section, we supplement the firing rate statistics of the NF-ResNet-34 model trained by OTTT$_A$ on ImageNet, as shown in Fig.~\ref{firing rate imagenet}. Overall the firing rate is around 0.24 and with 6 time steps each neuron averagely generate 1.46 spikes. Note that we can also reduce the time steps to realize a trade-off between accuracy and energy, as shown in Fig. 3 in Section 5.5. For example, with 2 time steps each neuron only averagely generate 0.48 spikes, with around 2.5\% accuracy drop.

\subsection{Comparison between OTTT and BPTT with Feedback Connections}

In this section, we supplement the results to compare the performance of OTTT and BPTT with feedback connections. As shown in Table~\ref{feedback ottt and bptt}, feedback connections can improve the performance for both OTTT and BPTT, and the improvement of OTTT from feedback connections is more significant than that of BPTT.

\begin{table} [ht]
	\centering
	\small
	\tabcolsep=2mm
	\captionof{table}{Performance on CIFAR-100 for VGG and VGG-F trained by OTTT$_O$ and BPTT.}
	\begin{tabular}{cccc}
		\toprule[1pt]
		Method & Network structure & Params & Mean$\pm$Std (Best)\\
		\midrule[0.5pt]
		OTTT$_O$ (ours) & VGG & 9.3M & 71.05$\pm$0.06\% (71.11\%)\\
		OTTT$_O$ (ours) & VGG-F & 9.6M & 72.63$\pm$0.23\% (72.94\%)\\
		\hline
		BPTT & VGG & 9.3M & 69.06$\pm$0.07\% (69.15\%)\\
		BPTT & VGG-F & 9.6M & (69.49\%)\\
		\bottomrule[1pt]
	\end{tabular}
	\label{feedback ottt and bptt}
\end{table}

\subsection{Experiments on Fully Recurrent Structures}

In this section, we supplement an experiment to use a recurrent spiking neural network to classify the Fashion-MNIST dataset~\cite{xiao2017fashion}. The input is flattened as a vector with 784 dimensions, and is connected to 400 spiking neurons with recurrent connections, which are then connected to a readout layer for classification. We apply weight standardization for connection weights from inputs to hidden neurons. Models are trained by 100 epochs with batch size 128 and SGD with momentum 0.9. The initial learning rate is set as 0.1 with a cosine annealing learning rate scheduler to 0. Dropout is set as 0.2, and weight decay is set as 5e-4 for BPTT and OTTT$_A$ while 1e-4 for OTTT$_O$ (since OTTT$_O$ update more times for each iteration). As for the loss function, we set $\alpha=0.05$ following CIFAR-10. As shown in Table~\ref{fashionmnist}, for this relatively simple model, the results of OTTT and BPTT are similar and BPTT performs slightly better.

\begin{table} [ht]
	\centering
	\small
	\tabcolsep=0.5mm
	\captionof{table}{Performance on Fashion-MNIST.}
	\begin{tabular}{cccc}
		\toprule[1pt]
		Method & Network structure & Time steps & Accuracy\\
		\midrule[0.5pt]
		ST-RSBP~\cite{zhang2019spike} & 400 (R400) & 400 & 90.00$\pm$0.14\% (90.13\%)\\
		IDE~\cite{xiao2021training} & 400 (R400) & 5 & 90.07$\pm$0.10\% (90.25\%) \\
		\hline
		BPTT & 400 (R400) & 5 & 90.58\% \\
		OTTT$_A$ (ours) & 400 (R400) & 5 & 90.36\%\\
		OTTT$_O$ (ours) & 400 (R400) & 5 & 90.40\%\\
		\bottomrule[1pt]
	\end{tabular}
	\label{fashionmnist}
\end{table}

\section{Discussion of Limitations and Social Impacts}

This work focus on online training of spiking neural networks, and therefore limits the usage of some techniques on network structures such as batch normalization along the temporal dimension. In this work, we adopt the scaled weight standardization as an alternative, which may require additional regularization to fully catch up the best performance of batch normalization as shown in the results of ANNs~\cite{brock2021characterizing}. It may require exploration of more techniques that is specific for SNNs to improve the performance and meanwhile compatible with more natural properties of SNNs, e.g. the online property.

As for social impacts, since this work focuses only on training methods for spiking neural networks, there is no direct negative social impact. And we believe that the development of successful energy-efficient SNN models could broader its applications and alleviate the huge energy consumption by ANNs. Besides, understanding and improving the training of biologically plausible SNNs may also contribute to the understanding of our brains and bridge the gap between biological neurons and successful deep learning.

\end{document}